\documentclass{article} 
\usepackage{iclr2027_conference,times}

\usepackage{amsmath,amsfonts,bm}

\def\eqref#1{equation~\ref{#1}}

\def\1{\bm{1}}

\def\vs{{\bm{s}}}

\DeclareMathAlphabet{\mathsfit}{\encodingdefault}{\sfdefault}{m}{sl}
\SetMathAlphabet{\mathsfit}{bold}{\encodingdefault}{\sfdefault}{bx}{n}

\usepackage{hyperref}
\usepackage{url}
\usepackage{booktabs}
\usepackage{multirow}
\usepackage{colortbl} 
\usepackage{makecell}
\usepackage{caption}
\usepackage{xspace}

\usepackage{pgfplots}
\pgfplotsset{compat=1.18}

\usepackage{soul}

\usepackage{lineno}
\usepackage{lineno}
\makeatletter
\newcommand*\patchAmsMathEnvironmentForLineno[1]{%
  \expandafter\let\csname old#1\expandafter\endcsname\csname #1\endcsname
  \expandafter\let\csname oldend#1\expandafter\endcsname\csname end#1\endcsname
  \renewenvironment{#1}%
     {\linenomath\csname old#1\endcsname}%
     {\csname oldend#1\endcsname\endlinenomath}}%
\makeatother

\usepackage{tcolorbox}
\tcbuselibrary{skins}
\usepackage{xcolor}

\title{Style over Substance? Linguistic Bias in LLM-based Peer Review}

\author{%
\textbf{Sai Suresh Macharla Vasu}$^{* 1}$,
\textbf{Sumedh Santosh Joshi}$^{* 1}$,
\textbf{Philipp Christmann}$^{1}$,
\textbf{Ivaxi Sheth}$^{1}$, \\
\textbf{Ruta Binkyte}$^{1,2}$,
\textbf{Koel Dutta Chowdhury}$^{3}$,
\textbf{Raouf Kerkouche}$^{4}$,
\textbf{Hui-Po Wang}$^{1}$,
\textbf{Mario Fritz}$^{1}$ \\[1.5ex]
$^{1}$CISPA Helmholtz Center for Information Security \\
$^{2}$German Research Center for Artificial Intelligence (DFKI) \\
$^{3}$University of Technology Nuremberg \\
$^{4}$INRIA, University of Lille, CNRS, Centrale Lille
}

\newcommand{\para}[1]{\noindent \textbf{#1}.}
\newcommand{\struct}[1]{\texttt{#1}}

\newcommand{\aspect}[1]{{\small \textsf{#1}}}
\newcommand{\criteria}[1]{{\struct{#1}}}
\newcommand{\llm}[1]{{\textsc{#1}}}

\definecolor{revisionblue}{HTML}{000000}
\newcommand{\revision}[1]{{\color{revisionblue}#1}}

\newcommand{\todo}[1]{}
\newcommand{\thought}[2]{}
\newcommand{\pc}[1]{}
\newcommand{\koel}[1]{}
\newcommand{\hpw}[1]{}

\definecolor{native}{HTML}{3B78C8}
\definecolor{nonnative}{HTML}{D9822B}
\definecolor{highconf}{HTML}{2E9D68}
\definecolor{lowconf}{HTML}{8B63C7}

\newcommand{\tmpnative}{\textcolor{native}{native}\xspace}
\newcommand{\tmpnonnative}{\textcolor{nonnative}{non-native}\xspace}
\newcommand{\tmphighconf}{\textcolor{highconf}{high-confidence}\xspace}
\newcommand{\tmplowconf}{\textcolor{lowconf}{low-confidence}\xspace}

\newcommand{\tabnative}{\textcolor{native}{\textbf{Native}}\xspace}
\newcommand{\tabnonnative}{\textcolor{nonnative}{\textbf{Non-native}}\xspace}
\newcommand{\tabhighconf}{\textcolor{highconf}{\textbf{High-confidence}}\xspace}
\newcommand{\tablowconf}{\textcolor{lowconf}{\textbf{Low-confidence}}\xspace}

\newcommand{\claudeopus}{\llm{Claude Opus 5}\xspace}

\newcommand{\gpt}{\llm{Gpt-5.5}\xspace}
\newcommand{\claude}{\llm{Claude Sonnet 4.6}\xspace}
\newcommand{\gemini}{\llm{Gemini 3.5 Flash}\xspace}
\newcommand{\gemma}{\llm{Gemma3}\xspace}
\newcommand{\qwen}{\llm{Qwen3 VL}\xspace}
\newcommand{\glm}{\llm{Glm4.7}\xspace}

\colorlet{nativebg}{native!15}
\colorlet{nonnativebg}{nonnative!15}
\colorlet{highconfbg}{highconf!15}
\colorlet{lowconfbg}{lowconf!15}

\newcommand{\native}[1]{%
    \begingroup
    \sethlcolor{nativebg}%
    \hl{#1}%
    \endgroup
}
\newcommand{\nonnative}[1]{%
    \begingroup
    \sethlcolor{nonnativebg}%
    \hl{#1}%
    \endgroup
}
\newcommand{\highconf}[1]{%
    \begingroup
    \sethlcolor{highconfbg}%
    \hl{#1}%
    \endgroup
}
\newcommand{\lowconf}[1]{%
    \begingroup
    \sethlcolor{lowconfbg}%
    \hl{#1}%
    \endgroup
}

\newcolumntype{H}{>{\setbox0=\hbox\bgroup}c<{\egroup}@{}}
\newcolumntype{G}{>{\columncolor [gray] {0.90}}c}

\newcommand{\squishlist}{
    \begin{list}{$\bullet$}{ 
        \setlength{\itemsep}{0pt}
        \setlength{\parsep}{1pt}
        \setlength{\topsep}{1pt}
        \setlength{\partopsep}{0pt}
        \setlength{\leftmargin}{1.5em}
        \setlength{\labelwidth}{1em}
        \setlength{\labelsep}{0.5em} 
    } 
}
\newcommand{\squishend}{
  \end{list}  }

\newcommand{\promptstart}{
    \small  
    \noindent
    \begin{tcolorbox}[
      title=Prompt,
      colback=gray!8,
      colframe=gray!50,
      colbacktitle=white,
      coltitle=black,
      fonttitle=\bfseries\small,
      arc=4pt,
      boxrule=0.6pt,
      attach boxed title to top left={yshift=-2mm, xshift=3mm},
      boxed title style={colback=white, colframe=gray!50, arc=3pt, boxrule=0.5pt}
    ]
}
\newcommand{\promptend}{
  \end{tcolorbox}
}

\iclrfinalcopy
\begin{document}

\maketitle
\begingroup
\renewcommand{\thefootnote}{*}
\footnotetext{Equal contribution.}
\endgroup
\setcounter{footnote}{0}

\begin{abstract}
The number of paper submissions to major AI conferences has increased
dramatically in recent years, substantially increasing the peer-review
workload of researchers.
At the same time, large language models (LLMs) are increasingly used to
support peer review, both by individual reviewers and by conference
organizations.
While LLM capabilities have improved considerably, their evaluations may
be influenced by linguistic characteristics that are unrelated to 
scientific content.
In this work, we introduce a controlled experimental framework for studying
such effects
by systematically perturbing the linguistic style of research papers.
Specifically, we rewrite papers to exhibit
(i) linguistic features associated with native or non-native English writing,
and
(ii) high- or low-confidence linguistic features.
To construct these interventions, we derive fine-grained rewriting
instructions grounded in the linguistic literature and apply them using LLMs.
Crucially, our interventions are designed to preserve the underlying scientific content.
We then evaluate the resulting papers using different LLM reviewers.
Our experiments expose a critical vulnerability in LLM-based peer review: reviewers systematically reward confident rhetoric independent of scientific content, 
raising serious concerns about the robustness of their assessments.
\end{abstract}

\section{Introduction}
\label{sec:intro}

The peer-review process is an integral part of scientific publishing, but also requires substantial time and effort from researchers.
This challenge has become increasingly pronounced with the growing number of submissions to major AI conferences.
In parallel, recent advances in large language models (LLMs) have opened up new opportunities to automate or support parts of the reviewing process~\citep{jin2024agentreview,lin2025breaking,dycke2026automatic}.
Individual reviewers use LLMs to assist with the review process, while LLMs are also increasingly seen as systematic tools for generating or supporting peer review~\citep{demetrio2025gen,baumann2026stop}.

However, the use of LLMs for peer review also introduces new risks.
LLM judgments can be influenced by properties that are unrelated to the actual quality of the evaluated content.
Prior work on LLM-as-a-judge has, for example, demonstrated biases related to the position of an answer or preferences towards outputs of the same LLM family~\citep{liu2024lost,wataoka2024self, ye2025justice}.
Similar concerns have been identified specifically for LLM-based peer review as well.
LLM reviewers can be susceptible to adversarial modifications of papers~\citep{lin2025breaking}, fail to reliably identify faulty scientific reasoning~\citep{dycke2026automatic}, and exhibit biases based on author-related information such as affiliation, seniority, or publication history~\citep{vasu2026justice}.
Moreover, recent results suggest that even paraphrasing or stylistic properties of a paper can substantially affect LLM-generated review scores~\citep{kaneko2026paraphrasing,kousha2026stylistic,baumann2026stop}. 
These findings raise the central question of our paper:
''\textit{How do we systematically measure whether LLM reviewers judge the research quality, rather than the linguistic style with which the research content is presented?}" 

In this work, we specifically study the impact of \emph{linguistic style} on LLM-based peer review.
Researchers have different linguistic backgrounds, and previous work has already found evidence that language proficiency and writing style can influence human peer review~\citep{politzer2020preliminary, saposnik2014effect, lee2006predictors, sayegh31writing}.
Motivated by these human studies,
we focus on two complementary dimensions that together signal \textit{epistemic authority}:
(i)~the \emph{language proficiency} reflected in the writing, and
(ii)~the \emph{confidence} with which the research is presented.
Language proficiency can signal \textit{author credibility}~\citep{lev2010don}, while confident linguistic framing can influence the perceived \textit{scientific credibility}~\citep{teigen2023dimensions}.
Both dimensions are
well characterized in the linguistic literature~\citep{hinkel2002second, hyland1996writing},
and should not substantially affect the assessment of the underlying research
quality.

\begin{figure*}
    \centering
    \includegraphics[width=0.9\textwidth]{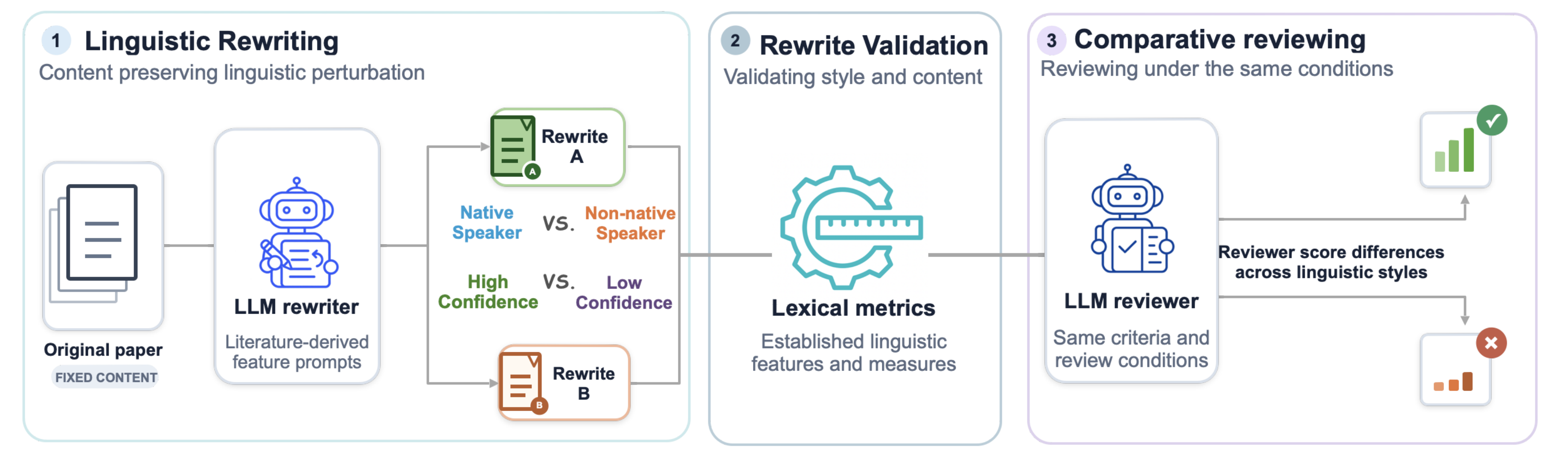}
    \vspace*{-0.4cm}
    \caption{Visualization of our controlled framework for linguistic rewriting of research papers.}
    \label{fig:overview}
\end{figure*}
Systematically studying such effects is challenging, as naturally occurring papers with different linguistic styles also differ in their research content and quality.
We therefore propose a controlled framework for perturbing a research paper along a specific linguistic dimension, while preserving its underlying research content.
Specifically, we rewrite papers to exhibit (i)~ linguistic characteristics associated with \tmpnative or \tmpnonnative English writing, and (ii)~\tmphighconf- or \tmplowconf linguistic characteristics.
Rather than relying on generic rewriting instructions, we derive fine-grained prompts from the linguistic literature to systematically control the corresponding stylistic features.
Crucially, we design these interventions to avoid distorting the content and thus the research quality of the paper, and verify the content integrity of the resulting rewrites.
To assess whether rewrites exhibit the intended stylistic characteristics
we use established linguistic measures.
The different rewrites are then given
to LLM reviewers, that provide us with a full-fledged peer review.
While review scores are the immediate quality signal, the arguments raised in reviews play an important role in the final decision process as well~\citep{arous2026hands}.
To this end, we integrate an automated annotation procedure for identifying positive and negative aspects in reviews~\citep{lu2025identifying}.

While related work has used counterfactual modifications to study reasoning capabilities~\citep{dycke2026automatic}, author-related biases~\citep{vasu2026justice}, or the general susceptibility of LLM reviewers to rewriting and paraphrasing~\citep{kaneko2026paraphrasing,baumann2026stop}, framework isolates linguistic cues of epistemic authority and examines their effects on review scores and qualitative arguments.
We use this framework to study different state-of-the-art LLM reviewers and investigate whether their reviews systematically favor perceived language proficiency and confidence.

Our experiments reveal that the two linguistic dimensions affect LLM reviewers differently. Across models, expressed confidence has a strong and consistent effect on evaluation even though the underlying scientific content remains unchanged.
In contrast, differences between \tmpnative- and \tmpnonnative associated writing are substantially smaller and more model-dependent.
More importantly, our controlled rewriting framework provides a practical way to stress-test LLM reviewers by varying specific linguistic cues and measuring when these changes alter review outcomes.
Our analysis further shows that these effects extend beyond review scores: LLM reviewers raise substantially different arguments across rewrites, focusing on different aspects of the paper and expressing different sentiments toward them.

In summary, our main contributions are:
\squishlist
\item We devise a controlled experimental framework for stress-testing LLM reviewers. By rewriting research papers along different linguistic dimensions in a content-preserving manner, our framework enables the systematic investigation of linguistic biases in LLM-based peer review.

\item We design literature-grounded rewriting prompts to manipulate language proficiency and expressed confidence, and validate the resulting rewrites using established linguistic measures.

\item We conduct experiments across state-of-the-art LLM reviewers. Our results reveal that expressed confidence has a strong and consistent effect on reviewer preferences, while effects associated with language proficiency are substantially smaller and vary across models.

\squishend

\section{Methodology}
\label{sec:exp-setup}

In this section, we introduce our experimental framework for
controlled linguistic interventions to detect stylistic biases in LLM reviewers.
Starting with the original paper, our framework proceeds in three stages.
First, we generate two versions that differ along a specific linguistic dimension, while preserving the underlying research content.
We then validate the content integrity of rewrites, and verify the presence of the respective linguistic style.
Finally, we provide the resulting versions to LLM reviewers and compare their reviews.
Figure~\ref{fig:overview} provides an overview of our framework.


\subsection{Linguistic Rewriting}
\label{sec:linguistic-rewriting}

A central challenge in designing the rewriting interventions is to obtain
linguistically distinguishable versions while changing only their linguistic realization and not the scientific content.
This challenge was also highlighted in prior work on linguistically informed LLM-rewriting \citep{kunilovskaya2024mitigating}.
A simple solution would be to directly ask an LLM to imitate a particular linguistic style. However, this could surface stereotypical characteristics~\citep{cheng2023marked} or perturb the scientific content itself, leading to incomparable rewrites.
Hence, 
we opt for a more principled approach and carefully design
rewriting prompts based on specific linguistic properties described in the corresponding literature.


\para{Preserving content integrity}
A crucial requirement of our interventions is that the scientific content, and thus the research quality of the paper, is preserved.
To this end, we design a set of \textit{content integrity rules} shared across rewriting prompts (Figures~\ref{fig:prompt-language-native} - \ref{fig:prompt-confidence-low}). These rules specify which content must remain unchanged and require preserving the length and structure of the original text.
At the same time, all rewrites must remain grammatical and plausible as 
professional
academic English, to avoid actually degrading the scientific quality by introducing grammatical mistakes or typos.



\para{Perturbing language proficiency}
Prior work identifies systematic differences between L1 and L2 English
academic writing at the lexical, phraseological, and syntactic levels.
\cite{crossley2009computational} identify differences
in lexical variation and sophistication, while
\citet{adel2012recurrent} find a larger and more varied repertoire of
recurrent lexical bundles in \tmpnative English writing.
\cite{hinkel2002second} further documents differences across a broader
set of lexical, syntactic, and rhetorical features.
Finally, L1 and L2 writers differ in the linguistic expressions they use to qualify claims and signal certainty, including their choice of words for conveying different degrees of epistemic commitment~\citep{hyland1997qualification}.

Based on these findings, we characterize language proficiency primarily
through the \emph{breadth and diversity of the available linguistic
repertoire}.
Our prompts therefore target lexical diversity and sophistication, the
repertoire of conventional academic expressions, syntactic variation, and
the range of epistemic expressions.
Importantly, lower proficiency is not represented through grammatical
errors or otherwise defective English.
Both rewriting directions represent plausible, publication-quality academic
writing and are designed to capture subtle tendencies reported in the
literature rather than stereotypical \tmpnative or \tmpnonnative writing.
The prompts are provided in
Figures~\ref{fig:prompt-language-native} and
\ref{fig:prompt-language-nonnative}.

\para{Perturbing expressed confidence}
Our second dimension captures the degree of expressed confidence expressed in
the writing.
A central mechanism for expressing epistemic commitment in academic writing
is the use of \emph{hedges} and \emph{boosters}: hedges qualify commitment
and introduce uncertainty, whereas boosters emphasize certainty and stronger
commitment~\citep{hyland1996writing,hyland1998boosting}.
Their use reflects different degrees of commitment and detachment in academic
discourse~\citep{vassileva2001commitment}, and more recent work similarly
studies epistemic positioning through such linguistic
markers~\citep{yang2025find}.

Hence, we perturb papers along the confidence dimension by varying the degree of epistemic
commitment expressed through hedging, boosting, modality, and qualification.
The \tmphighconf and \tmplowconf prompts move these properties in opposite
directions while avoiding exaggerated certainty or implausible hesitation.
A central challenge when designing the prompts was to distinguish \emph{expressed confidence} from the \emph{substantive strength} of a scientific claim: its linguistic formulation may become more or less cautious, but its scope, evidence, and underlying
scientific meaning must remain unchanged.
The rewriting prompts are provided in
Figures~\ref{fig:prompt-confidence-high} and
\ref{fig:prompt-confidence-low}.

Table~\ref{tab:rewrite-examples} provides example rewrites for both dimensions along with the original version.

\begin{table*}[t]
    \centering
    \scriptsize
    \begin{tabular}{p{0.14\textwidth} p{0.82\textwidth}}
        \toprule
        \textbf{Rewrite} & \textbf{Example} \\
        \midrule
        \textbf{Original}
        &
        Systematically studying such effects is challenging, as naturally
        occurring papers with different linguistic styles also differ in their
        research content and quality. We therefore propose a controlled
        experimental framework for perturbing a research paper along a
        specific linguistic dimension, while preserving its underlying research
        content.
        \\

        \midrule

        \textbf{\textcolor{native}{Native}}
        &
        Systematically \native{investigating} such effects is \native{no
        straightforward matter}, since naturally occurring papers written in
        \native{divergent} linguistic styles also tend to differ in their
        research content and quality. We therefore \native{put forward} a
        controlled experimental framework for perturbing a research paper
        along one specific linguistic dimension while \native{leaving} its
        underlying research content \native{intact}.
        \\

        \vspace{0.05cm}
        \textbf{\textcolor{nonnative}{Non-native}}
        &
        \vspace{0.05cm}
        \nonnative{To study} such effects \nonnative{in a systematic way} is
        \nonnative{difficult}, because naturally occurring papers that differ in
        linguistic style also differ in their research content and quality. We
        therefore propose a controlled experimental framework \nonnative{which
        perturbs} a research paper along one specific linguistic
        dimension, while the underlying research content is preserved.
        \\

        \midrule

        \textbf{\textcolor{highconf}{High-confidence}}
        &
        Systematically studying such effects \highconf{poses a clear
        methodological challenge}, since naturally occurring papers with
        different linguistic styles \highconf{invariably} differ in their
        research content and quality as well. We therefore \highconf{introduce}
        a controlled experimental framework that perturbs a research paper
        along a specific linguistic dimension while preserving its underlying
        research content.
        \\

        \vspace{0.05cm}
        \textbf{\textcolor{lowconf}{Low-confidence}}
        &
        \vspace{0.05cm}
        A systematic study of such effects \lowconf{appears} challenging,
        \lowconf{in part} because naturally occurring papers differing in
        linguistic style \lowconf{tend also to} differ in their research content
        and quality. We therefore propose a controlled experimental framework
        \lowconf{intended to} perturb a research paper along a specific
        linguistic dimension, while \lowconf{seeking to} preserve its underlying
        research content.
        \\

        \bottomrule
    \end{tabular}
    \vspace*{-0.2cm}
    \caption{Example rewrites of the same scientific passage, taken from our own
    introduction, along the two linguistic dimensions. Colored spans highlight
    selected linguistic changes induced by the language-proficiency and
    expressed-confidence interventions. Further examples from our methodology
    and results sections are provided in
    Appendix~\ref{sec:more-rewrite-examples}.}
    \label{tab:rewrite-examples}
\end{table*}


\subsection{Rewrite Validation}
\label{sec:rewrite-validation}

The validity of our experimental setup depends on two central properties of
the generated rewrites.
First, the rewrite should preserve the scientific content of the original
paper.
Second, the rewriting process should systematically induce the intended
linguistic characteristics.
We therefore evaluate both \emph{content integrity} and \emph{linguistic change} for each rewriting dimension.

\para{Validating the content integrity}
First of all, we validate that our systematic rewrites preserve the content integrity of the original paper, and are thus comparable.
To this end, we consider two different measures: \textit{semantic similarity} and \textit{lexical preservation}. 
Specifically, we compute the semantic similarity between each rewrite and the original paper using dense embeddings.\footnote{We use \href{https://huggingface.co/BAAI/bge-large-en-v1.5}{\llm{BAAI/bge-large-en-v1.5}} for embedding passages.}
For long documents, comparing dense embeddings directly can overestimate similarity scores~\citep{xiao2023length}.
Hence, each paper is split into passages, and we then compute the average maximum cosine similarity between original and rewritten passage embeddings.
This ensures that we capture content integrity on a more fine-grained level than with pure paper- or section-level embeddings.
In addition to semantic similarity, we quantify the lexical preservation using cosine similarity between unigram and bigram TF-IDF representations for the original paper and rewrites.
This metric complements semantic similarity and validates whether the technical terminology is preserved.

We report the content-integrity scores in Table~\ref{tab:rewrite-validation}.
Across all rewrites, both measures strongly indicate the preservation of scientific content, with scores close to their maximum value of $1$.

\revision{
We further assess factual consistency using AlignScore~\citep{zha-etal-2023-alignscore}
and sentence-level preservation across all 504 papers.
AlignScore ranges from 0.895 to 0.982 across rewrites,
while 57.1--80.2\% of aligned sentences remain identical
to the originals.
Removing hedging and boosting expressions
increases the low-confidence AlignScore from 0.895 to 0.935.
These additional checks provide further evidence of
content preservation (Appendix~\ref{sec:additional-content-integrity}).
}

\para{Validating language proficiency perturbations}
We validate the language-proficiency rewrites using established
measures for lexical and textual complexity.
First, we use \emph{Measure of Textual Lexical Diversity} (MTLD) to
measure lexical diversity~\citep{mccarthy2010mtld}.
This directly reflects our instruction to vary the breadth and
diversity of the vocabulary.
Second, we use \emph{Advanced Guiraud} (AG) to measure lexical sophistication~\citep{daller2003lexical}, 
capturing the use of advanced vocabulary.
Both measures are motivated by prior work for identifying systematic differences in lexical variation and sophistication between L1 and L2 English
writing~\citep{crossley2009computational}.
Finally, we report the Gunning Fog index (GF) as a broader measure of textual complexity~\citep{gunning1952technique}, which considers sentence length and the prevalence of complex words.

Table~\ref{tab:rewrite-validation} shows the results for this validation step.
The linguistic measures demonstrate the viability of our rewriting prompts for perturbing the language proficiency: all metrics show an increase for \tmpnative-associated rewrites compared to \tmpnonnative-associated rewrites.

\para{Validating expressed confidence perturbations}
For expressed confidence, we directly measure the linguistic properties targeted
by our rewriting prompts, and measure the frequency of \emph{boosters} and \emph{hedges} per 1000 words.

As shown in Table~\ref{tab:rewrite-validation},
\tmphighconf rewrites contain substantially more boosters and fewer hedges than \tmplowconf rewrites, confirming the successful manipulation of expressed confidence.

\subsection{Comparative Reviewing}
\label{sec:llm-reviewer}

After systematically rewriting papers and validating the underlying linguistic properties,
we can stress-test LLM reviewers along different linguistic dimensions.
Our framework enables to determine whether papers of the same research quality receive systematically different evaluations based on their linguistic realization.
For each paper and rewriting dimension, the LLM reviewer receives one rewritten version at a time and produces a complete review together with an overall review score.
We then compare the scores assigned to the corresponding versions of the same original paper.

\para{Review instructions}
We derive a reviewer prompt from standard reviewing instructions at major AI conferences, including explicit review criteria and guidance on the expected output format.
The specific prompt is provided in Figure~\ref{fig:prompt-review}.
We conducted preliminary experiments with different prompts and found the LLM reviewers to be largely insensitive to such prompt variations: the direction and significance of all effects remain unchanged
(details in Appendix~\ref{sec:prompt-sensitivity}).

\para{Review output}
The reviewer output contains the paper's strengths,
weaknesses, and suggestions for improvement.
In addition, it includes scores for \emph{soundness},
\emph{significance}, \emph{evidence}, and \emph{clarity} on a 1--4 scale, a reviewer confidence score on a 1--5 scale, and an overall rating on a 1--10 scale.
We explicitly outline the Likert scale for the overall rating in the prompt to avoid any ambiguities.

\subsection{Configuration}
\label{sec:configuration}

\para{Research papers}
We sample $504$ research papers from the 2026 reviewing cycle of ICLR, uniformly across the $21$ primary areas. From each primary area, we randomly sample $12$ accepted and $12$ rejected papers.
All papers are anonymized, simulating a double blind peer-review setup.

\para{Rewriting model}
For paper rewriting, we use \claudeopus in all experiments to ensure strong adherence to our prompts.
Keeping the rewriting model fixed prevents differences across reviewers from being confounded by variation in the rewriting model.

\para{LLM reviewers}
For reviewing, we consider LLMs from different model families and capability levels.
We choose only LLMs with a knowledge cutoff before 2026, to avoid any confounding factors due to known acceptance decisions for ICLR 2026.
As proprietary models, we select \gpt, \claude and \gemini, as they reflect the default model choices in the respective consumer-facing chat interface as of that cutoff.
In addition, we consider the open-weight models 
\gemma (27B), \qwen (235B) and \glm (358B).

\para{Reviewer variance}
Since LLM-reviewer scores can vary across repeated generations, even for identical inputs, we quantify the reviewer variance for every reviewer-model to help interpret our results.
We randomly sample $10$ papers and conduct $n$=$10$ independent repeated reviews for each LLM. The variance for each model is reported in Table~\ref{tab:reviewer-variance}.
Reviewer variance differs substantially across models: \gpt, \claude, \gemma, and \qwen exhibit pooled standard deviations of approximately $0.5$ rating points, while \glm ($0.85$) and particularly \gemini ($1.24$) show greater variability.
This enables us to contextualize score differences between linguistic rewrites relative to the inherent stochastic variation of the corresponding reviewer.

We provide further technical details, including LLM generation parameters, in Appendix~\ref{sec:impl-details}.

\begin{table*}[t]
    \centering
    \resizebox*{\textwidth}{!}{
    \begin{tabular}{l ccc Hc Hc ccc}
        \toprule
        \textbf{Rewrite} &
        \multicolumn{3}{c}{\textbf{Language Proficiency}} &
        \multicolumn{4}{c}{\textbf{Expressed Confidence}} &
        \multicolumn{3}{c}{\textbf{Content Integrity}} \\
        \cmidrule(lr){2-4}
        \cmidrule(lr){5-8}
        \cmidrule(lr){9-11}
        & \textbf{MTLD}
        & \textbf{AG}
        & \textbf{GF}
        & \textbf{\%chunks}
        & \textbf{Boosters/1K words}
        & \textbf{\%chunks}
        & \textbf{Hedges/1K words}
        & \textbf{Semantic}
        & \textbf{TF-IDF}
        & \revision{\textbf{AlignScore}} \\
        \midrule

        \textbf{Original}
            & 60.8
            & 7.7
            & 17.0
            & 11.9
            & 1.9
            & 9.0
            & 1.5
            & --
            & --
            & \revision{--} \\
        \midrule

        \tabnative
            & \native{63.9}
            & \native{8.4}
            & \native{17.4}
            & 11.3
            & 1.7
            & 13.9
            & 2.3
            & 0.997
            & 0.992
            & \revision{0.963} \\

        \tabnonnative
            & 58.0
            & 7.2
            & 16.8
            & 10.5
            & 1.6
            & 9.4
            & 1.5
            & 0.998
            & 0.995
            & \revision{0.982} \\

        \midrule

        \tabhighconf
            & 60.8
            & 7.8
            & 17.2
            & 18.0
            & \highconf{3.2}
            & 6.2
            & 0.8
            & 0.999
            & 0.998
            & \revision{0.965} \\

        \tablowconf
            & 62.9
            & 7.7
            & 17.2
            & 3.9
            & 0.7
            & 38.7
            & \lowconf{13.6}
            & 0.998
            & 0.993
            & \revision{0.895} \\

        \bottomrule
    \end{tabular}
    }
    \vspace*{-0.2cm}
    \caption{
        Validation of the language-proficiency and confidence rewrites.
        For language proficiency, we report lexical diversity (\textbf{MTLD}),
        lexical sophistication (\textbf{AG}), and textual complexity (\textbf{GF}).
        For confidence, we report the average frequency of boosters and hedges per
        1,000 words. For both rewrite dimensions, we additionally report
        our measures for content integrity based on {semantic similarity} (\textbf{Semantic}) and lexical similarity (\textbf{TF-IDF}) to the original paper.
        \revision{We additionally report factual consistency using
        \textbf{AlignScore}.}
        Highlighted values mark the higher value within each pair of rewrites,
        for the measures targeted by the respective rewrite dimension.
    }
    \label{tab:rewrite-validation}
\end{table*}

\begin{table}[t]
    \centering
    \scriptsize
    \begin{tabular}{H cccccc}
        \toprule
        Model
        & \textbf{\gpt}
        & \textbf{\claude}
        & \textbf{\gemini}
        & \textbf{\gemma}
        & \textbf{\qwen}
        & \textbf{\glm}
        \\
        
        \midrule
        \textbf{SD}
        & 0.48
        & 0.49
        & 1.24
        & 0.52
        & 0.49
        & 0.85 \\
        \bottomrule
    \end{tabular}
    \vspace*{-0.3cm}
    \caption{Pooled within-paper standard deviation of reviewer scores
    across repeated generations.}
    \label{tab:reviewer-variance}
\end{table}

\section{Experimental Results}
\label{sec:exp-results}

\begin{table*}[t]
    \centering
    \small
    \resizebox*{\textwidth}{!}{
        \begin{tabular}{llllc GG c}
        \toprule
        & & & & & \multicolumn{2}{c}{\textbf{Decision Flips (\%)}} &  \\
        \cmidrule(lr){6-7}
        \textbf{Comparison} & \textbf{Reviewer}  &
        \textbf{Preference (\%)} & \textbf{Ratio} & \textbf{$p$} & \textbf{First $\uparrow$} & \textbf{Second $\uparrow$}
        & \textbf{Average Scores} \\
        \midrule

        \multirow{6}{*}{
            \makecell[l]{%
                \tabnative\\
                vs.\ \tabnonnative
            }
        }
            & \gpt    & \native{17} / 14 (69)    & \native{1.3$\times$}    & 0.15 & 6.2  & 3.0  & 3.6 / 3.6 \\
            & \claude & 19 / \nonnative{21} (60) & \nonnative{0.9$\times$} & 0.67 & 7.5  & 8.5  & 4.5 / 4.5 \\
            & \gemini & \native{33} / 32 (36)    & {1.0$\times$}           & 0.78 & 11.1 & 10.9 & 4.7 / 4.7 \\
            \cmidrule(lr){2-8}
            & \gemma  & \native{24} / 19 (57)    & \native{1.2$\times$}    & 0.13 & 0.2  & 0.0  & 7.2 / 7.2 \\
            & \qwen   & \native{19} / 17 (63)    & \native{1.1$\times$}    & 0.51 & 0.0  & 0.0  & 8.0 / 8.0 \\
            & \glm    & \native{37} / 34 (29)    & \native{1.1$\times$}    & 0.46 & 12.0 & 12.2 & \native{6.2} / 6.1 \\

        \midrule

        \multirow{6}{*}{
            \makecell[l]{%
                \tabhighconf\\
                vs.\ \tablowconf
            }
        }
            & \gpt    & \highconf{20} / 8 (72)  & \highconf{2.7$\times$} & \textbf{$<$0.001} & 5.6  & 1.0  & \highconf{3.6} / 3.5 \\
            & \claude & \highconf{40} / 6 (54)  & \highconf{7.0$\times$} & \textbf{$<$0.001} & 21.0 & 2.8  & \highconf{4.6} / 4.2 \\
            & \gemini & \highconf{36} / 27 (37) & \highconf{1.3$\times$} & \textbf{0.01}     & 15.1 & 10.1 & \highconf{4.7} / 4.5 \\
            \cmidrule(lr){2-8}
            & \gemma  & \highconf{52} / 9 (39)  & \highconf{6.0$\times$} & \textbf{$<$0.001} & 0.2  & 0.0  & \highconf{7.2} / 6.7 \\
            & \qwen   & \highconf{36} / 10 (55) & \highconf{3.7$\times$} & \textbf{$<$0.001} & 0.2  & 0.0  & \highconf{8.0} / 7.7 \\
            & \glm    & \highconf{55} / 19 (26) & \highconf{2.9$\times$} & \textbf{$<$0.001} & 19.6 & 7.9  & \highconf{6.1} / 5.3 \\

        \bottomrule
        \end{tabular}
    }
    \vspace*{-0.2cm}
    \caption{
        Pairwise reviewer preferences for \tmpnative vs.\ \tmpnonnative and
        \tmphighconf vs.\ \tmplowconf rewrites (504 papers).
        \textbf{Preference}: \% of papers where the first / second version
        scores higher (ties in parentheses); \textbf{Ratio}: first-to-second version preference ratio;
        \textbf{$p$}: two-sided sign test, significant values ($p<0.05$) in bold.
        \textbf{Decision Flips}: \% of papers where only the first / second version is accepted.
        \textbf{Average Scores}: mean overall rating of the first / second
        version.
        Highlighted values indicate the preferred direction.
    }
    \label{tab:merged_preferences}
\end{table*}

\subsection{Main Results}
\label{sec:main-res}

We first investigate whether linguistic rewrites
systematically affect the scores assigned by LLM reviewers.
Our primary comparisons are \textit{\tmpnative vs.\ \tmpnonnative} for language proficiency and \textit{\tmphighconf vs.\ \tmplowconf} for expressed confidence.
For each pair, we report the percentage of the first version receiving a higher review score, the second version receiving a higher score, or both versions receiving the same score (i.e., a tie).
Table~\ref{tab:merged_preferences} reports our main results.

\para{Preferences along the linguistic dimensions}
The first takeaway is that the two linguistic dimensions affect reviewer scores differently.
For the language proficiency dimension, preferences are comparatively weak and inconsistent across reviewers. For 5/6 of the reviewers, \tmpnative rewrites receive higher scores more often, but the corresponding preference ratios remain subtle (between $1.0\times$ and $1.3\times$).
In contrast, we observe a consistent and substantially stronger preference towards \tmphighconf over \tmplowconf rewrites across all LLM reviewers.~{\textcolor{highconf}{High-confidence}} rewrites are preferred between $1.3\times$ and $7.0\times$ as often as their \tmplowconf counterparts. The most notable preference ratios are observed for \claude ($7.0\times$) and \gemma ($6.0\times$).

The average scores show a similar pattern: scores are nearly identical between \tmpnative and \tmpnonnative rewrites, but consistently higher for \tmphighconf than for \tmplowconf rewrites.

\para{Impact on review decisions}
\label{sec:decision-flips}
A change in review score becomes more significant as it flips the acceptance decision of the paper. We therefore analyze how frequently our linguistic interventions result in such decision flips. Specifically, we define a \emph{decision flip} whenever the two rewrites of the same paper fall on opposite sides of the acceptance threshold (4.5 in our case).
Table~\ref{tab:merged_preferences} reports both directions separately:
\tmpnonnative rejection $\rightarrow$ \tmpnative acceptance (\textbf{First$\uparrow$})
and \tmpnative rejection $\rightarrow$ \tmpnonnative acceptance (\textbf{Second$\uparrow$}),
and for confidence \tmplowconf rejection $\rightarrow$ \tmphighconf acceptance (\textbf{First$\uparrow$}) and the reverse direction (\textbf{Second$\uparrow$}).

For language proficiency, decision flips are largely balanced, which is consistent with the absence of a systematic score preference.
For confidence, flips predominantly favor the \tmphighconf version.
We observe the most drastic effect for \claude: in $21.0\%$ of reviews, \claude accepts the \tmphighconf version and reject the \tmplowconf version, with only $2.8\%$ of papers accepted only for the \tmplowconf rewrite.
As the average scores for \gemma and \qwen lie well above the acceptance threshold, decision flips are rare for these models.


\subsection{In-depth Analysis}
\label{sec:analysis}

\para{Combined effects of linguistic style}
\label{sec:combined-effects}
Our main experiments consider language proficiency and confidence in isolation.
This analysis investigates how both dimensions interact when they occur simultaneously.
For a batch of 126 papers, we generate four additional rewrites,
including all combinations of the two dimensions.
The corresponding rewriting prompts are merged (see Figures~\ref{fig:prompt-interaction-native_high}-\ref{fig:prompt-interaction-nonnative_low} in the Appendix).
We then investigate
(i) the language proficiency dimension given a specific confidence rewrite,
(ii) the confidence dimension given a specific proficiency rewrite,
(iii) the comparison of the two extremes (\tmpnative\unskip\&\tmphighconf vs. \tmpnonnative\unskip\&\tmplowconf),
and (iv) the inverse (\tmpnative\unskip\&\tmplowconf vs.\ \tmpnonnative\unskip\&\tmphighconf).

\revision{

Table~\ref{tab:all_pairwise_comparisons_frontier}
shows the results across all 6 reviewers.
The conditional comparisons largely reproduce the effects
observed when varying the dimensions in isolation.
Holding confidence fixed, differences between \tmpnative
and \tmpnonnative rewrites remain relatively small and
mostly non-significant, except for \gpt under high
confidence ($p=0.017$).
The preference for \tmphighconf rewrites persists when
combined with proficiency, significantly for \claude, \gemma, \qwen, and \glm reviewers under both conditions,
and for \gpt under native proficiency.
The cross-condition comparisons further illustrate the
strength of the confidence effect.
Combining \tmpnative and \tmphighconf language yields
significant preferences over \tmpnonnative\unskip\&
\tmplowconf rewrites for five reviewers, with preference
ratios of up to $9.1\times$.
Reversing confidence switches this pattern, with all
reviewers favoring \tmpnonnative but \tmphighconf rewrites,
significantly so for \claude, \gemma, \qwen, and \glm.
Overall, confidence dominates these comparisons,
although \gemini shows no significant effects.

\begin{table*}[t]
    \centering
    \scriptsize
    \setlength{\tabcolsep}{4pt}
    \begin{tabular}{llllcc}
        \toprule
        \textbf{Comparison} & \textbf{Condition} & \textbf{Model} &
        \textbf{Preference (\%)} & \textbf{Ratio} & \textbf{$p$} \\
        \midrule

        \multirow{12}{*}{%
            \makecell[l]{%
                \tabnative\\
                vs. \tabnonnative
            }}
        & \multirow{6}{*}{Given \tabhighconf}
            & \gpt    & \native{13} / 4 (83)   & \native{3.4$\times$} & \textbf{0.017} \\
        &   & \claude & 13 / \nonnative{15} (72) & \nonnative{0.9$\times$} & 0.851 \\
        &   & \gemini & \native{32} / 31 (37) & \native{1.0$\times$} & 1.000 \\
        \cmidrule(lr){3-6}
        &   & \gemma & \native{27} / 19 (54) & \native{1.4$\times$} & 0.24 \\
        &   & \qwen  & \native{20} / 17 (63) & \native{1.2$\times$} & 0.66 \\
        &   & \glm   & \native{39} / 29 (33) & \native{1.4$\times$} & 0.19 \\

        \cmidrule(lr){2-6}

        & \multirow{6}{*}{Given \tablowconf}
            & \gpt    & \native{14} / 10 (75) & \native{1.4$\times$} & 0.473 \\
        &   & \claude & \native{19} / 9 (72) & \native{2.1$\times$} & 0.087 \\
        &   & \gemini & 28 / 28 (44) & 1.0$\times$ & 1.000 \\
        \cmidrule(lr){3-6}
        &   & \gemma & \native{32} / 19 (49) & \native{1.7$\times$} & 0.06 \\
        &   & \qwen  & 17 / \nonnative{25} (58) & \nonnative{0.7$\times$} & 0.27 \\
        &   & \glm   & \native{37} / 30 (33) & \native{1.2$\times$} & 0.45 \\

        \midrule

        \multirow{12}{*}{%
            \makecell[l]{%
                \tabhighconf\\
                vs. \tablowconf
            }}
        & \multirow{6}{*}{Given \tabnative}
            & \gpt    & \highconf{18} / 8 (74) & \highconf{2.3$\times$} & \textbf{0.035} \\
        &   & \claude & \highconf{27} / 9 (64) & \highconf{3.0$\times$} & \textbf{0.004} \\
        &   & \gemini & \highconf{34} / 29 (36) & \highconf{1.2$\times$} & 0.609 \\
        \cmidrule(lr){3-6}
        &   & \gemma & \highconf{43} / 10 (47) & \highconf{4.2$\times$} & \textbf{$<$0.001} \\
        &   & \qwen  & \highconf{37} / 9 (54) & \highconf{4.3$\times$} & \textbf{$<$0.001} \\
        &   & \glm   & \highconf{48} / 27 (25) & \highconf{1.8$\times$} & \textbf{0.007} \\

        \cmidrule(lr){2-6}

        & \multirow{6}{*}{Given \tabnonnative}
            & \gpt    & \highconf{14} / 9 (77) & \highconf{1.6$\times$} & 0.265 \\
        &   & \claude & \highconf{33} / 5 (62) & \highconf{6.6$\times$} & \textbf{$<$0.001} \\
        &   & \gemini & 30 / 30 (41) & 1.0$\times$ & 1.000 \\
        \cmidrule(lr){3-6}
        &   & \gemma & \highconf{45} / 8 (47) & \highconf{5.7$\times$} & \textbf{$<$0.001} \\
        &   & \qwen  & \highconf{30} / 10 (60) & \highconf{2.9$\times$} & \textbf{$<$0.001} \\
        &   & \glm   & \highconf{53} / 21 (25) & \highconf{2.5$\times$} & \textbf{$<$0.001} \\

        \midrule

        \multirow{6}{*}{%
            \makecell[l]{%
                \tabnative\& \tabhighconf\\
                vs. \tabnonnative\& \tablowconf
            }}
        & & \gpt    & \native{22} / 7 (71) & \native{3.0$\times$} & \textbf{0.004} \\
        & & \claude & \native{31} / 5 (64) & \native{6.2$\times$} & \textbf{$<$0.001} \\
        & & \gemini & \native{33} / 29 (38) & \native{1.1$\times$} & 0.699 \\
        \cmidrule(lr){3-6}
        & & \gemma & \native{51} / 6 (44) & \native{9.1$\times$} & \textbf{$<$0.001} \\
        & & \qwen  & \native{31} / 11 (58) & \native{2.8$\times$} & \textbf{$<$0.001} \\
        & & \glm   & \native{55} / 22 (23) & \native{2.5$\times$} & \textbf{$<$0.001} \\

        \midrule

        \multirow{6}{*}{%
            \makecell[l]{%
                \tabnative\& \tablowconf\\
                vs. \tabnonnative\& \tabhighconf
            }}
        & & \gpt    & 10 / \nonnative{12} (78) & \nonnative{0.9$\times$} & 0.851 \\
        & & \claude & 4 / \nonnative{24} (72) & \nonnative{0.2$\times$} & \textbf{$<$0.001} \\
        & & \gemini & 25 / \nonnative{39} (36) & \nonnative{0.6$\times$} & 0.124 \\
        \cmidrule(lr){3-6}
        & & \gemma & 12 / \nonnative{37} (51) & \nonnative{0.3$\times$} & \textbf{$<$0.001} \\
        & & \qwen  & 7 / \nonnative{33} (60) & \nonnative{0.2$\times$} & \textbf{$<$0.001} \\
        & & \glm   & 27 / \nonnative{48} (25) & \nonnative{0.6$\times$} & \textbf{0.007} \\

        \bottomrule
    \end{tabular}
    \vspace*{-0.2cm}
    \caption{
        \revision{
        Pairwise reviewer preferences across combined
        linguistic rewrites (126 papers).
        \textbf{Preference}: \% of papers where the first /
        second version scores higher (ties in parentheses);
        \textbf{Ratio}: first-to-second preference ratio;
        \textbf{$p$}: two-sided sign test, with significant
        values ($p<0.05$) in bold.
        Highlighted values indicate the preferred direction.
        }
    }
    \label{tab:all_pairwise_comparisons_frontier}
\end{table*}

}
    
\para{Preferences along the language proficiency dimension}
\label{sec:biased-reviewer}
Our main results discover substantial effects of linguistic rewrites along the confidence dimension,
but only subtle changes along the proficiency dimension.
A natural question that arises is whether LLM reviewers are indeed not substantially affected by language proficiency, or whether our rewrites are not sufficient for bringing out existing biases (despite validated rewriting effects by linguistic metrics; see Table~\ref{tab:rewrite-validation}).
To answer this question, we simulate biased reviewers by injecting a dedicated instruction into the reviewer prompt:
``\textit{In addition to assessing the scientific content, place substantial weight on the linguistic style and overall presentation of the paper \dots}'' 
(full prompt in Figure~\ref{fig:prompt-review-biased}).
We intentionally keep the prompt general, and avoid mentioning language proficiency directly.
We run this analysis on 126 papers, equally drawn from the primary areas (see Table~\ref{tab:biased-reviewer}
in the Appendix for the full results).

Our framework reveals the simulated biased reviewers, and identifies strong preference ratios towards \tmpnative-associated writing of \native{6.8} (\gpt), \native{4.3} (\claude),  and \native{2.2} (\gemini),
all of which are statistically significant under a two-sided sign
test ($p<0.001$ for \gpt and \claude, $p<0.01$ for \gemini).
These results demonstrate that our framework can reveal both the \textit{presence} (this experiment) and \textit{absence} (Table~\ref{tab:merged_preferences}) of systematic preferences along stylistic dimensions, supporting its utility for stress-testing LLM reviewers.

\begin{figure}[t]
  \centering
    \includegraphics[width=0.7\textwidth]{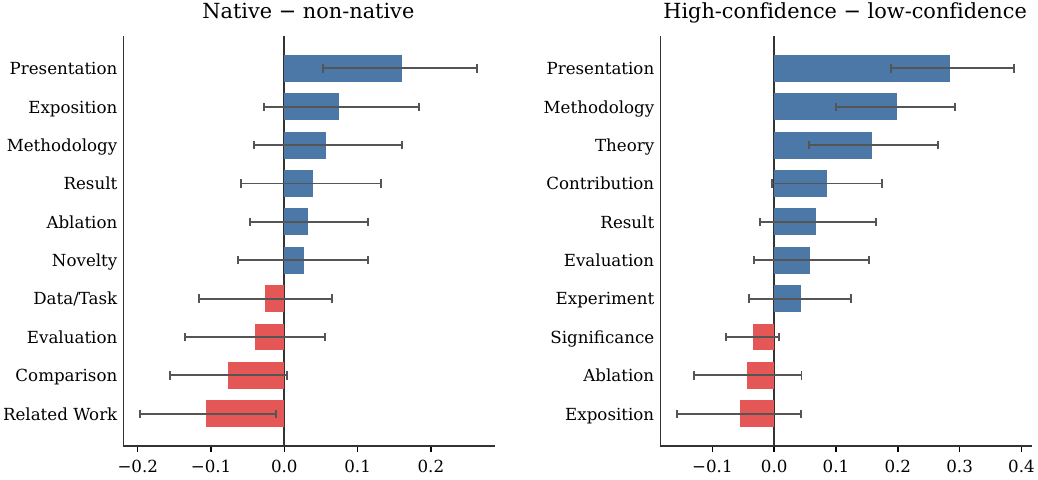}
    \vspace*{-0.3cm}
    \caption{Mean aspect-level review polarity. For each paper and aspect, we score review texts as \#positive mention -- \#negative mentions, compute within-paper difference between rewrites and average over all papers. 
    Left: \tmpnative\unskip$-$\tmpnonnative. Right: \tmphighconf\unskip$-$\tmplowconf.
    Bars show the top-10 aspects with largest $|\Delta|$. Error bars are 95\% bootstrap CIs.
    }
    \label{fig:review-aspects}
\end{figure}

\revision{
\para{Comparison with original papers}
To contextualize the effects of our linguistic interventions,
we additionally evaluate all 504 original, unmodified papers
using the same six LLM reviewers.
Scores for \tmplowconf rewrites are significantly lower than for originals
across all six models ($p \leq 0.013$).
Further, the \tmphighconf rewrites score significantly higher than the
originals for \gpt ($p = 0.011$) and \gemma ($p < 0.001$),
showing that more confident writing can increase review scores.
The full results are reported in
Appendix~\ref{sec:original-baseline}.

\para{Targeted confidence ablations}
Our confidence rewrites simultaneously vary hedges and boosters,
making it difficult to attribute the observed effects to either
linguistic feature.
To investigate their individual contributions, we conduct three
targeted ablations on all 504 papers using \gpt, \claude, and \gemini.
We align corresponding sentences and selectively modify hedging
and boosting expressions while preserving the surrounding text.

Adding hedges to \tmphighconf rewrites significantly reduces review
scores across all three models ($p < 0.001$), while removing boosters has a smaller effect,
with significant effects only for \claude.
Conversely, adding boosters to \tmplowconf does not necessarily increase review
scores, but their effect depends on the linguistic context.
Detailed results and manipulation checks are provided in
Appendix~\ref{sec:confidence-ablation}.

\para{Accepted versus rejected papers}
To investigate whether linguistic preferences depend on the
underlying research quality, we stratify the 504 papers by
acceptance status (252 accepted and 252 rejected).
The \tmphighconf rewrites receive significantly higher scores
than \tmplowconf rewrites in both groups across nearly
all six reviewers, with the sole exception of \gemini
on rejected papers ($p = 0.316$).
In contrast, preferences between \tmpnative and \tmpnonnative
rewrites are not statistically significant in either group.
Detailed results are provided in
Appendix~\ref{sec:acceptance-stratified}.

\para{Alignment with human reviewers}
We compare LLM reviewer scores with human evaluations
on 504 original ICLR 2026 papers.
\claude achieves the highest alignment (0.882),
exceeding the human--human baseline (0.835), and
distinguishes accepted from rejected papers (AUROC = 0.71).
After cross-validated Platt scaling, its Brier score
is 0.215, compared to 0.250 for a constant-probability baseline.
These results indicate meaningful human alignment without
ruling out linguistic biases
(Appendix~\ref{sec:human-alignment}).

}

\para{Aspects that drive the reviews}
\label{sec:review-focus}
Review scores alone do not decide the outcome of the peer review process. Arguments brought up in the review are also decisive in deriving an acceptance decision~\citep{arous2026hands}.
We therefore extend our study beyond the plain review scores, and investigate which aspects of a paper~\citep{lu2025identifying} drive the reviewers' assessments and whether this focus changes with linguistic style.
In particular, we study whether linguistic variations affect the attention
given to the underlying scientific content.

We utilize the taxonomy established by \cite{lu2025identifying}, and consider the $16$ coarse-grained aspects they defined, including aspects such as \aspect{Presentation}, \aspect{Methodology}, \aspect{Novelty} or \aspect{Related Work}. The annotation is automated via an LLM\footnote{We use \href{https://huggingface.co/google/gemma-4-26B-A4B-it}{\llm{Gemma4 (26B)}} for these simple annotations.}, which provides a list of aspects with their connotation (positive or negative). In addition, the annotator provides the exact phrase the aspect was derived from,
which we verify via substring matching (all of the derived aspects were valid).
The prompt we used for aspect annotations is provided in Figure~\ref{fig:prompt-aspect annotation} in the Appendix.


We then investigate the aspect-level polarity under linguistic rewriting. 
For each review, aspect and polarity, we count positive and negative mentions of aspects, and define the net polarity of an aspect as \#positive mention -- \#negative mentions. Along one dimension, we then compute the within-paper difference between rewrites (\tmpnative--\tmpnonnative and \tmphighconf--\tmplowconf), and average this difference over all papers.
The results are provided in Figure~\ref{fig:review-aspects}.

While proficiency rewrites have only subtle effects on review scores,
our aspect-level analysis reveal differences in how the LLM reviewer perceive the rewrites:
for \tmpnative-associated papers \aspect{Presentation} is less often criticized ($\Delta$=$-2.4$\,pp) and more often praised ($\Delta$=$+3.0$\,pp), leading to a net-polarity increase of $0.16$.
Interestingly, this behavior is similar for the confidence dimension, where \aspect{Presentation} also has the largest net-polarity increase ($+0.29$).
These shifts in perceived \aspect{Presentation} quality do not systematically affect review scores in both cases ($r$=$0.04$),
which is a desired property in this case.

Finally, we find that the overall rating strongly correlates with the share of positive aspect mentions ($r$=$0.81$).
For \tmphighconf rewrites, aspects such as \aspect{Methodology}, \aspect{Theory} and \aspect{Contribution} are perceived more positively, alongside the larger \aspect{Presentation} gain, contributing to the increase in positive mention share,
and thus contributing to the preference ratios observed in Tables~\ref{tab:merged_preferences}~and~\ref{tab:all_pairwise_comparisons}.

\section{Related Work}
\label{sec:related}

\para{LLMs for peer review}
With increasing paper submissions, conferences and reviewers have explored the use of LLMs to assist with peer review \citep{jin2024agentreview,demetrio2025gen,thakkar2026large, zhu2025deepreview, huang2025papereval}. While promising, LLM-based reviewing raises reliability and fairness concerns, including biases toward author-related properties such as affiliation, seniority, and publication history~\citep{ye2024we, zhu2025deepreview, vasu2026justice}, as well as sensitivity to textual adversarial attacks and prompt injections. Most closely related, \citet{kousha2026stylistic} find that linguistic complexity and text length are more strongly associated with ChatGPT-based research evaluations than with expert-based evaluations in many fields. We complement this work by applying controlled, linguistically grounded transformations to the same papers, manipulating language proficiency and expressed confidence while preserving research content. This isolates effects of linguistic presentation from differences across papers.

\para{Linguistic biases in human peer review}
Prior linguistic research documents systematic differences between L1 and L2 English writing. \citet{crossley2009computational} identify differences in lexical variation and sophistication, while \citet{adel2012recurrent} find that native English writers employ a larger and more varied repertoire of recurrent lexical bundles. Differences also extend to epistemic expression: \citet{hyland1997qualification} show that L1 and L2 writers differ in the range and precision with which they express qualification and certainty. Linguistic presentation can also affect human peer review. In a controlled experiment, \citet{politzer2020preliminary} find preliminary evidence that identical scientific content receives lower ratings when presented in non-native-like rather than native-like English, although the manual interventions limited the study to paper abstracts. Similarly, an observational study of submissions to \emph{Stroke} finds lower priority scores and acceptance rates for submissions from non-English-speaking countries, even after controlling for potential confounders~\citep{saposnik2014effect}. Beyond language proficiency, \citet{sayegh31writing} find a modest association between writing style and publication recommendations in a simulated peer-review experiment. These findings motivate our controlled investigation of whether linguistic presentation similarly affects LLM-based peer review, particularly given evidence of linguistic biases in other LLM applications~\citep{hofmann2024ai, reusens2025native, deas2023evaluation}.

\section{Conclusion}
\label{sec:discussion}

In this work, we propose a controlled framework for investigating linguistic biases in LLM-based peer review.
Our framework applies interventions to research papers along linguistic markers of epistemic authority, while preserving the underlying scientific content.
Our experiments reveal substantial preferences of LLM reviewers toward rewrites using \tmphighconf language.
Crucially, these preferences frequently result in decision flips across the accept$\leftrightarrow$reject boundary, directly affecting review outcomes.
Beyond scores, we find that aspects raised in reviews vary in frequency and sentiment depending on linguistic features of the paper, potentially influencing final decisions.

Our findings raise important concerns for the use of LLMs in peer review. 
If content-equivalent papers receive different scores and review arguments solely because of linguistic properties, reviewer judgments do not exclusively reflect scientific quality.
LLM reviewers should therefore be stress-tested on content-preserving rewrites to determine whether linguistic changes systematically affect their assessments.
These tests should cover multiple linguistic dimensions and examine both overall scores and review arguments.
In particular, reviewer systems should be checked for whether expressed confidence is treated as a signal of stronger evidence or scientific quality.
Our framework enables these tests and comparisons of reviewer configurations by the stability of their assessments under controlled linguistic interventions.
Finally, substantial differences in average scores and within-model variance caution against treating raw scores from individual LLM reviewers as directly comparable without calibration.
Future work can extend these stress tests to additional linguistic and rhetorical dimensions.

\subsection*{AI use statement}

In this work, we used generative AI tools to support the design and refinement of the experiments, to implement and edit software code, and to generate the LLM-based reviews and linguistic rewrites that constitute experimental data in our study.
We additionally used generative AI tools for analyzing existing literature, identifying relevant literature, and editing the manuscript to improve readability and presentation.
We did not use generative AI tools to formulate mathematical claims or assist in writing mathematical proofs, and tasks involving surveys, interviews, or transcription of research material are not applicable to this work.
All AI-assisted research and writing activities were reviewed by the authors. We take responsibility for the final content of this work, including text, claims, code, experimental artifacts, and analyses produced with the aid of generative AI.




\subsection*{Ethics statement}


This work does not advocate the use of LLMs as replacements for human peer reviewers. Instead, we develop a controlled framework to identify and characterize biases in LLM-based reviewing. Our study shows that linguistic properties of a paper can affect LLM-generated review outcomes even when the underlying research content is preserved. Such sensitivity raises fairness concerns, as linguistic variation associated with researchers' backgrounds may influence how their work is evaluated.

Our findings also have potential for dual use. Identifying linguistic features that systematically influence LLM reviewers could be used to deliberately manipulate papers in order to obtain more favorable automated reviews. We do not intend to facilitate such manipulation. Rather, we use these findings to expose potential vulnerabilities of LLM-based reviewing and to motivate their careful evaluation before such systems are used in high-stakes peer-review settings.
\subsection*{Reproducibility Statement}

All rewriting prompts are provided verbatim in
Figures~\ref{fig:prompt-language-native}--\ref{fig:prompt-interaction-nonnative_low},
and the reviewer prompts in Figures~\ref{fig:prompt-review} and
\ref{fig:prompt-review-biased}.
The paper processing pipeline, the rewriting and reviewing models, and all
generation settings are described in Section~\ref{sec:configuration} and
Appendix~\ref{sec:impl-details}; the variations of the reviewer prompt used in
our robustness analysis are described in Appendix~\ref{sec:prompt-sensitivity}.
The sampling of the 504 ICLR 2026 papers is described in
Section~\ref{sec:configuration}, and the rewrite validation measures in
Section~\ref{sec:rewrite-validation}.
Since several reviewer models do not support fixed sampling parameters, we
quantify the variance of each reviewer across repeated generations
(Table~\ref{tab:reviewer-variance}).
We will release our code, the list of sampled papers, all rewrites and all
generated reviews upon publication.

\bibliography{iclr2027_conference}
\bibliographystyle{iclr2027_conference}

\clearpage
\appendix

\section{Implementation Details}
\label{sec:impl-details}

\para{Paper processing}
We convert each paper from PDF to a structured textual representation using
MinerU~(v3.4.4), which preserves the document layout so that tables and
equations remain inline at their original positions. Before conversion, we cut
each PDF after the references, so that no appendix reaches any model. We then
split the paper into top-level sections and blind it, removing the title block
with authors and affiliations, the venue watermark, page numbers and the
line-number gutter. Figures are not passed to any model; their captions are
kept as text.

To preserve the overall organization of the paper and avoid exceeding model
context or output limits, we perform rewriting section by section.
When rewriting a section, the model receives no context, to ensure rewrites
are only stylistic and no additional content is added.

\para{Confidence Indicators} To quantify expressed confidence, we measure the use of boosters (e.g., “clearly,” “demonstrates,” “certainly”) and hedges (e.g., ``may'' ``suggests'', ``likely'') drawn from Hyland’s metadiscourse taxonomy~\citep{hyland2005metadiscourse}. Each paper was segmented into chunks, normalized (lowercased, punctuation-stripped), and scanned for booster and hedge terms using word-boundary-anchored matching (``\textbackslash{b}'' in regular expressions) to ensure only whole-word occurrences were counted. We report frequency per 1,000 words for each category, with booster and hedge terms kept mutually exclusive to avoid double-counting.

\para{Generation settings}
All rewrites are produced with Claude Opus~5. For all rewriting and reviewing
experiments, we use each model's default sampling configuration; we do not set
the temperature, since several of the models used do not accept a custom
value, and fixing it for some models but not others would make the reviewers
less comparable. For GLM~4.7, we explicitly enable reasoning; all other models
run with their default reasoning behaviour. Each review is generated in a
single call, with the system prompt in Figure~\ref{fig:prompt-review} and the
blinded paper in Markdown as the user message.
We use the same generation settings across conditions.
\revision{

\section{Additional Content-Integrity Validation}
\label{sec:additional-content-integrity}

We complement the semantic and lexical similarity measures
reported in the main paper (Section~\ref{sec:rewrite-validation} with two additional analyses.

\para{Factual consistency}
We use AlignScore~\citep{zha-etal-2023-alignscore} to evaluate
the factual consistency of each rewrite against its
original paper.
Our implementation tests both directions (original$\rightarrow$rewrite and rewrite$\rightarrow$original),
and we then provide the mean AlignScore for both directions.

\para{Sentence preservation}
We align sentences between each original paper and its
rewrites and calculate the percentage of aligned sentences
that remain identical. We repeat both evaluations after
removing hedging and boosting expressions (no Hedges/Boosters) to
examine their contribution to the measured differences.

\begin{table}[t]
    \centering
    \small
    \setlength{\tabcolsep}{4pt}
    \begin{tabular}{lcccc}
        \toprule
        \textbf{Rewrite} &
        \makecell{\textbf{Align}\\\textbf{Score}} &
        \makecell{\textbf{AlignScore}\\\textbf{(no Hedges/Boosters)}} &
        \makecell{\textbf{Identical}\\\textbf{(\%)}} &
        \makecell{\textbf{Identical}\\\textbf{(no Hedges/Boosters, \%)}} \\
        \midrule
        High-confidence & 0.965 & 0.965 & 80.2 & 81.6 \\
        Low-confidence  & 0.895 & 0.935 & 71.9 & 73.9 \\
        Native          & 0.963 & 0.960 & 57.1 & 57.3 \\
        Non-native      & 0.982 & 0.980 & 66.9 & 67.2 \\
        \bottomrule
    \end{tabular}
    \caption{\revision{Factual consistency and sentence preservation
    across linguistic rewrites (504 papers). No H/B denotes
    measurements after removing hedging and boosting
    expressions.}}
    \label{tab:additional-content-integrity}
\end{table}

\para{Results}
Table~\ref{tab:additional-content-integrity} presents
the factual consistency and sentence-preservation results.
AlignScore remains high across all rewrites
(0.895--0.982). Removing hedging and boosting expressions
increases the low-confidence AlignScore from 0.895 to
0.935, demonstrating that AlignScore is quite sensitive to such hedges and boosters.

Furthermore, 57.1--80.2\% of aligned sentences remain
identical to the originals. Together with the semantic
and lexical similarity measures reported in the main
paper, these results provide additional evidence that
our rewrites largely preserve the underlying scientific
content.
}
\section{Further Rewrite Examples}
\label{sec:more-rewrite-examples}

Tables~\ref{tab:rewrite-examples-method} and~\ref{tab:rewrite-examples-results}
show rewrites of two further passages from this paper, taken from the
methodology and the results section, produced in the same way as
Table~\ref{tab:rewrite-examples}.

\begin{table*}[t]
    \centering
    \scriptsize
    \begin{tabular}{p{0.14\textwidth} p{0.82\textwidth}}
        \toprule
        \textbf{Rewrite} & \textbf{Example} \\
        \midrule
        \textbf{Original}
        &
        The validity of our experimental setup depends on two central
        properties of the generated rewrites. First, the rewrite should
        preserve the scientific content of the original paper. Second, the
        rewriting process should systematically induce the intended linguistic
        characteristics. We therefore evaluate both content integrity and
        linguistic change for each rewriting dimension.
        \\
        \midrule
        \textbf{\textcolor{native}{Native}}
        &
        The validity of our experimental setup \native{rests on} two central
        properties of the generated rewrites. First, a rewrite \native{ought to}
        preserve the scientific content of the original paper. Second, the
        rewriting \native{procedure} should \native{reliably} induce the
        intended linguistic characteristics. We therefore \native{assess} both
        content integrity and linguistic change for each rewriting dimension.
        \\
        \vspace{0.05cm}
        \textbf{\textcolor{nonnative}{Non-native}}
        &
        \vspace{0.05cm}
        The validity of our experimental setup \nonnative{is based on} two
        \nonnative{main} properties of the generated rewrites. First, the
        rewrite should \nonnative{keep} the scientific content of the original
        paper. Second, the rewriting process should induce the intended
        linguistic characteristics \nonnative{in a systematic way}.
        \nonnative{For this reason}, we evaluate both content integrity and
        linguistic change for each rewriting dimension.

        \\
        \midrule
        \textbf{\textcolor{highconf}{High-confidence}}
        &
        The validity of our experimental setup rests on two central properties
        of the generated rewrites. First, the rewrite \highconf{preserves} the
        scientific content of the original paper. Second, the rewriting process
        \highconf{systematically induces} the intended linguistic
        characteristics. We therefore evaluate both content integrity and
        linguistic change for each rewriting dimension.
        \\
        \vspace{0.05cm}
        \textbf{\textcolor{lowconf}{Low-confidence}}
        &
        \vspace{0.05cm}
        The validity of our experimental setup \lowconf{appears to rest on} two
        properties of the generated rewrites. First, the rewrite should,
        \lowconf{as far as possible}, preserve the scientific content of the
        original paper. Second, the rewriting process should \lowconf{plausibly}
        induce the intended linguistic characteristics \lowconf{in a reasonably
        systematic manner}. We therefore \lowconf{attempt to} assess both
        content integrity and linguistic change for each rewriting dimension.
        \\
        \bottomrule
    \end{tabular}
    \vspace*{-0.1cm}
 \caption{Example rewrites of a passage from our methodology section along
    the two linguistic dimensions. Colored spans highlight selected linguistic
    changes induced by the language-proficiency and expressed-confidence
    interventions.}
    \label{tab:rewrite-examples-method}
\end{table*}

\begin{table*}[t]
    \centering
    \scriptsize
    \begin{tabular}{p{0.14\textwidth} p{0.82\textwidth}}
        \toprule
        \textbf{Rewrite} & \textbf{Example} \\
        \midrule
        \textbf{Original}
        &
        For language proficiency, decision flips are largely balanced, which is
        consistent with the absence of a systematic score preference. For
        confidence, flips predominantly favor the high-confidence version. We
        observe the most drastic effect for \claude: in 21.0\% of reviews,
        \claude accepts the high-confidence version and reject the
        low-confidence version, with only 2.8\% of papers accepted only for the
        low-confidence rewrite.
        \\
        \midrule
        \textbf{\textcolor{native}{Native}}
        &
        \native{With respect to} language proficiency, decision flips are
        \native{for the most part} balanced, \native{a pattern consistent with}
        the absence of any systematic score preference. For confidence,
        \native{by contrast}, flips predominantly favor the high-confidence
        version. The \native{most pronounced} effect \native{emerges} for
        \claude: in 21.0\% of reviews, \claude accepts the high-confidence
        version \native{while rejecting} the low-confidence one,
        \native{whereas a mere} 2.8\% of papers are accepted \native{solely} in
        their low-confidence rewrite.
        \\
        \vspace{0.05cm}
        \textbf{\textcolor{nonnative}{Non-native}}
        &
        \vspace{0.05cm}
        For language proficiency, decision flips are \nonnative{mostly}
        balanced, which is \nonnative{in line with} the absence of a systematic
        score preference. For confidence, flips \nonnative{mostly} favor the
        high-confidence version. The most drastic effect \nonnative{is observed}
        for \claude: in 21.0\% of reviews, \claude accepts the high-confidence
        version and rejects the low-confidence version, \nonnative{and only}
        2.8\% of papers are accepted only for the low-confidence rewrite.
        \\
        \midrule
        \textbf{\textcolor{highconf}{High-confidence}}
        &
        For language proficiency, decision flips are largely balanced, which is
        \highconf{fully} consistent with the absence of a systematic score
        preference. For confidence, flips \highconf{clearly} favor the
        high-confidence version. The most drastic effect \highconf{occurs} for
        \claude: in 21.0\% of reviews, \claude accepts the high-confidence
        version and rejects the low-confidence version, with only 2.8\% of
        papers accepted only for the low-confidence rewrite.
        \\
        \vspace{0.05cm}
        \textbf{\textcolor{lowconf}{Low-confidence}}
        &
        \vspace{0.05cm}
        For language proficiency, decision flips \lowconf{appear to be} largely
        balanced, which \lowconf{would seem} consistent with the absence of a
        systematic score preference. For confidence, flips \lowconf{tend to}
        favor the high-confidence version. The most drastic effect
        \lowconf{appears to occur} for \claude: in 21.0\% of reviews, \claude
        accepts the high-confidence version and rejects the low-confidence
        version, with only 2.8\% of papers accepted only for the low-confidence
        rewrite.
        \\
        \bottomrule
    \end{tabular}
    \vspace*{-0.1cm}
    \caption{Example rewrites of a passage from our results section along the
    two linguistic dimensions. Colored spans highlight selected linguistic
    changes induced by the language-proficiency and expressed-confidence
    interventions.}
    \label{tab:rewrite-examples-results}
\end{table*}

\section{Prompt Sensitivity}
\label{sec:prompt-sensitivity}

To test whether our findings depend on the specific reviewer prompt, we re-ran
\claude on Batch~1 (126 papers) with three neutral variations of our
prompt: \textit{Brief} (without the reviewer responsibilities, review criteria
and rating-scale anchors), \textit{Minimal} (additionally without the sentence
on fair and substantive feedback), and \textit{Bare} (additionally without the
four per-criterion sub-scores). As a contrast, we also ran a \textit{Biased}
prompt that explicitly instructs the reviewer to place substantial weight on
linguistic style and presentation.
Default refers to the standard prompt used throughout this work (Figure~\ref{fig:prompt-review}).

Table~\ref{tab:prompt-sensitivity} shows that the neutral variations leave our
conclusions unchanged: the \tmpnative vs.\ \tmpnonnative contrast remains a null effect
under every neutral prompt, and the high- vs.\ low-confidence contrast remains
large and significant ($p<0.0001$). Only the explicitly biased prompt changes
the picture, turning the \tmpnative vs.\ \tmpnonnative null result into a clear
preference for \tmpnative writing (4.3$\times$, $p<0.0001$). Reviewers are thus
insensitive to how the review task is worded, but do act on an explicit
instruction to weight style.

\begin{table}[t]
    \centering
    \small
    \begin{tabular}{ll ccc cc c}
        \toprule
        & & & & & \multicolumn{2}{c}{\textbf{Decision Flips (\%)}} & \\
        \cmidrule(lr){6-7}
        \textbf{Comparison} & \textbf{Prompt} & \textbf{Preference (\%)} & \textbf{Ratio} & \textbf{$p$}
        & \textbf{First $\uparrow$} & \textbf{Second $\uparrow$} & \textbf{Avg.\ Scores} \\
        \midrule
        \multirow{5}{*}{\makecell[l]{\tabnative\\vs.\ \tabnonnative}}
            & Default & 17 / \nonnative{21} (62) & \nonnative{0.8$\times$} & 0.47 & 7.1 & 7.9 & 4.3 / 4.3 \\
            & Brief   & \native{19} / 14 (67)    & \native{1.3$\times$}    & 0.44 & 7.9 & 4.8 & \native{4.5} / 4.4 \\
            & Minimal & 15 / \nonnative{17} (67) & \nonnative{0.9$\times$} & 0.76 & 7.1 & 4.8 & 4.6 / 4.6 \\
            & Bare    & 14 / 14 (71)             & {1.0$\times$}           & 1.00 & 3.2 & 4.8 & 4.7 / 4.7 \\
            \cmidrule(lr){2-8}
            & Biased  & \native{44} / 10 (45)    & \native{4.3$\times$}    & \textbf{$<$0.001} & 19.0 & 4.0 & \native{4.8} / 4.4 \\
        \midrule
        \multirow{4}{*}{\makecell[l]{\tabhighconf\\vs.\ \tablowconf}}
            & Default & \highconf{36} / 4 (60) & \highconf{9.0$\times$} & \textbf{$<$0.001} & 18.3 & 3.2 & \highconf{4.3} / 3.9 \\
            & Brief   & \highconf{35} / 4 (61) & \highconf{8.8$\times$} & \textbf{$<$0.001} & 17.5 & 3.2 & \highconf{4.5} / 4.1 \\
            & Minimal & \highconf{41} / 6 (53) & \highconf{7.4$\times$} & \textbf{$<$0.001} & 26.2 & 2.4 & \highconf{4.6} / 4.2 \\
            \cmidrule(lr){2-8}
            & Biased & \highconf{27} / 7 (66) & \highconf{4.0$\times$} & \textbf{$<$0.001} & 14.5 & 0.9 & \highconf{4.4} / 4.1 \\
        \bottomrule
    \end{tabular}
    \caption{Prompt sensitivity of \claude on a subset of 126 papers.
    Columns as in Table~\ref{tab:merged_preferences}; significant $p$-values
    ($p<0.05$) in bold. Rows below the rule use the Biased prompt, which
    explicitly instructs the reviewer to weight linguistic style.}
    \label{tab:prompt-sensitivity}
\end{table}

\begin{table*}[t]
    \centering
    \small
    \resizebox*{\textwidth}{!}{
        \begin{tabular}{llll cc c}
        \toprule
        & & & & \multicolumn{2}{c}{\textbf{Decision Flips (\%)}} &  \\
        \cmidrule(lr){5-6}
        \textbf{Prompt} & \textbf{Reviewer} &
        \textbf{Preference (\%)} & \textbf{Ratio} & \textbf{First $\uparrow$} & \textbf{Second $\uparrow$}
        & \textbf{Average Scores} \\
        \midrule

        \multirow{6}{*}{\textbf{Default}}
            & \gpt    & \native{16} / 13 (71) & \native{1.2$\times$}   & 3.2  & 4.0  & 3.5 / 3.5 \\
            & \claude & 17 / \nonnative{21} (62) & \nonnative{0.8$\times$} & 7.1 & 7.9 & 4.3 / 4.3 \\
            & \gemini & \native{34} / 30 (36) & \native{1.1$\times$}   & 9.5  & 11.9 & \native{4.8} / 4.7 \\
            \cmidrule(lr){2-7}
            & \gemma  & \native{31} / 19 (50) & \native{1.6$\times$}   & 0.8  & 0.0  & \native{7.2} / 7.1 \\
            & \qwen   & \native{19} / 13 (67) & \native{1.4$\times$}   & 0.0  & 0.0  & \native{7.9} / 7.8 \\
            & \glm    & \native{38} / 35 (27) & \native{1.1$\times$}   & 14.3 & 12.7 & \native{6.1} / 5.9 \\
        \midrule

        \multirow{6}{*}{\textbf{Biased}}
            & \gpt    & \native{27} / 4 (69)  & \native{6.8$\times^\dagger$} & 14.3 & 0.8 & \native{3.9} / 3.6 \\
            & \claude & \native{44} / 10 (45) & \native{4.3$\times^\dagger$} & 19.0 & 4.0 & \native{4.8} / 4.4 \\
            & \gemini & \native{35} / 16 (49) & \native{2.2$\times^\dagger$} & 4.8  & 2.4 & \native{6.5} / 6.3 \\
            \cmidrule(lr){2-7}
            & \gemma  & \native{16} / 6 (79)  & \native{2.9$\times^\dagger$} & 0.0  & 0.0 & \native{8.1} / 8.0 \\
            & \qwen   & \native{28} / 17 (56) & \native{1.7$\times$}         & 0.0  & 0.0 & \native{8.4} / 8.3 \\
            & \glm    & 37 / 37 (26)          & {1.0$\times$}                & 9.1  & 6.6 & 6.6 / 6.6 \\
        \bottomrule
        \end{tabular}
    }
    \vspace*{-0.2cm}
    \caption{
        \tmpnative vs.\ \tmpnonnative preferences under the Default and the
        Biased reviewer prompt (Figure~\ref{fig:prompt-review-biased}), on the
        same 126 papers. Columns as in Table~\ref{tab:merged_preferences};
        $^\dagger$ marks ratios that are statistically significant
        (two-sided sign test, $p<0.05$).
    }
    \label{tab:biased-reviewer}
\end{table*}

\revision{
\section{Targeted Confidence Ablations}
\label{sec:confidence-ablation}

Our main confidence comparison varies hedges and boosters together. To
examine their individual contributions, we conduct targeted ablations.
Specifically, we align corresponding sentences and apply deterministic, targeted edits to the existing
confidence rewrites.
Our implementation is based on dictionaries, derived from Hyland’s metadiscourse taxonomy~\citep{hyland2005metadiscourse}, and simple rules and heuristics.
This yields three additional variants:
\begin{itemize}
    \item \textbf{High-confidence without added boosters:} Remove the
    boosters introduced by the high-confidence rewrite while retaining
    its low level of hedging.
    \item \textbf{High-confidence with added hedges:} Add hedges to the
    high-confidence rewrite while retaining its boosters.
    \item \textbf{Low-confidence with added boosters:} Add boosters to
    the low-confidence rewrite while largely retaining its high level
    of hedging.
\end{itemize}
Each variant is compared with the corresponding full confidence rewrite
of the same paper, evaluated by the same reviewer model.

\para{Manipulation checks}
We measure hedge and booster frequencies per 1,000 words.
Table~\ref{tab:confidence-markers} confirms that the
targeted edits change the intended linguistic markers
while largely preserving the respective other marker.

\begin{table}[t]
    \centering
    \small
    \setlength{\tabcolsep}{4pt}
    \begin{tabular}{lcc}
        \toprule
        \textbf{Rewrite} &
        \makecell{\textbf{Hedges}\\\textbf{/1K words}} &
        \makecell{\textbf{Boosters}\\\textbf{/1K words}} \\
        \midrule
        \tabhighconf          & 0.7  & 3.2 \\
        \tabhighconf w/o boosters   & 0.7  & 1.4 \\
        \tabhighconf w/ hedges      & 7.0  & 3.3 \\
        \tablowconf w/ boosters     & 12.3 & 2.9 \\
        \tablowconf           & 13.6 & 0.8 \\
        \bottomrule
    \end{tabular}
    \caption{\revision{Hedge and booster frequencies across
    confidence rewrites and targeted ablations
    (504 papers).}}
    \label{tab:confidence-markers}
\end{table}

\begin{table}[t]
    \centering
    \small
    \begin{tabular}{llcc}
        \toprule
        \textbf{Comparison} & \textbf{Reviewer} &
        \textbf{Preference (\%)} & \textbf{$p$} \\
        \midrule
        \multirow{3}{*}{\makecell[l]{\tabhighconf without boosters\\vs. \tabhighconf }}
          & \gpt    & 14 / 15 (71) & 0.560 \\
          & \claude & 14 / 23 (63) & \textbf{0.001} \\
          & \gemini & 28 / 31 (41) & 0.485 \\
        \midrule
        \multirow{3}{*}{\makecell[l]{\tabhighconf with hedges\\vs. \tabhighconf. }}
          & \gpt    & 7 / 23 (70)  & \textbf{\boldmath$<0.001$} \\
          & \claude & 6 / 39 (55)  & \textbf{\boldmath$<0.001$} \\
          & \gemini & 26 / 40 (34) & \textbf{\boldmath$<0.001$} \\
        \midrule
        \multirow{3}{*}{\makecell[l]{\tablowconf with boosters\\vs. \tablowconf }}
          & \gpt    & 7 / 21 (72)  & \textbf{\boldmath$<0.001$} \\
          & \claude & 12 / 23 (65) & \textbf{\boldmath$<0.001$} \\
          & \gemini & 22 / 37 (41) & \textbf{\boldmath$<0.001$} \\
        \bottomrule
    \end{tabular}
    \vspace*{-0.2cm}
    \caption{\revision{Pairwise preferences under targeted hedge/booster ablations
    (504 papers). \textbf{Preference}: percentage of papers where the
    first / second version receives a higher overall review score;
    ties are shown in parentheses. \textbf{$p$}: two-sided sign test;
    significant values ($p<0.05$) are in bold.}}
    \label{tab:confidence-ablation}
\end{table}

\para{Results}
We conduct experiments on
all 504 papers using \gpt, \claude, and \gemini as LLM reviewers.
Adding hedges to high-confidence rewrites lowers review scores across
all three models ($p<0.001$). Removing boosters from high-confidence
rewrites has a significant effect only for \claude ($p=0.001$), which
prefers the full high-confidence version. Adding boosters to
low-confidence rewrites also leads to lower scores than the unchanged
low-confidence versions across all three models ($p<0.001$).
Together, these ablations show a consistent penalty for increased
hedging, while adding boosters does not necessarily increase review
scores, but their effect depends on the linguistic context.
}
\revision{
\section{Comparison with Original Papers}
\label{sec:original-baseline}

Our main experiments compare high- and low-confidence rewrites of the
same paper, enabled by our controlled experimental framework.
To establish how these rewrites compare with the authors'
unmodified writing, we additionally evaluate all 504 original ICLR 2026
papers using the same six LLM reviewers. We compare each original paper
with its corresponding high- and low-confidence rewrites.

\begin{table*}[t]
    \centering
    \small
    \resizebox{\textwidth}{!}{%
    \begin{tabular}{lccclcc}
        \toprule
        \textbf{Reviewer} & \textbf{Original} & \textbf{\tabhighconf} &
        \textbf{\tablowconf} & \textbf{Score ordering} &
        \makecell{\textbf{High vs. Orig.}\\\textbf{($p$)}} &
        \makecell{\textbf{Low vs. Orig.}\\\textbf{($p$)}} \\
        \midrule
        \llm{GPT-5.5}            & 3.55 & 3.62 & 3.48 & High $>$ Original $>$ Low & \textbf{0.011} & \textbf{0.013} \\
        \llm{Claude Sonnet 4.6} & 4.68 & 4.58 & 4.17 & Original $>$ High $>$ Low & \textbf{\boldmath$<0.001$} & \textbf{\boldmath$<0.001$} \\
        \llm{Gemini 3.5 Flash}  & 5.29 & 4.74 & 4.52 & Original $>$ High $>$ Low & \textbf{\boldmath$<0.001$} & \textbf{\boldmath$<0.001$} \\
        \llm{Gemma 3 27B}       & 7.12 & 7.24 & 6.73 & High $>$ Original $>$ Low & \textbf{\boldmath$<0.001$} & \textbf{\boldmath$<0.001$} \\
        \llm{Qwen3 VL 235B}     & 8.02 & 8.02 & 7.71 & High $\approx$ Original $>$ Low & 0.944 & \textbf{\boldmath$<0.001$} \\
        \llm{GLM 4.7}           & 6.28 & 6.12 & 5.34 & Original $\approx$ High $>$ Low & 0.052 & \textbf{\boldmath$<0.001$} \\
        \bottomrule
    \end{tabular}%
    }
    \vspace*{-0.2cm}
    \caption{\revision{Comparison of original papers with confidence rewrites
    (504 papers; mean overall review scores on a 1--10 scale).
    High and Low denote the high- and low-confidence rewrites;
    Original denotes the unmodified paper. The ordering summarizes
    the observed mean scores; $\approx$ marks a non-significant
    High--Original comparison, not statistical equivalence.
    The $p$-values are from two-sided paired sign tests; significant
    values ($p<0.05$) are in bold.}}
    \label{tab:original-baseline}
\end{table*}

\para{Results}
As shown in Table~\ref{tab:original-baseline}, low-confidence rewrites
receive significantly lower scores than the originals across all six
reviewers ($p\leq0.013$). High-confidence rewrites score significantly
above the originals for \gpt ($p=0.011$) and \gemma ($p<0.001$),
showing that more confident writing can increase review scores.
In contrast, \claude and \gemini score the originals significantly
higher, while \qwen and \glm show no significant difference between
the originals and high-confidence rewrites.

Note that the original papers are not a style-neutral condition:
their use of hedges and boosters varies naturally.
These comparisons complement the direct high- versus
low-confidence results and the targeted ablations in
Appendix~\ref{sec:confidence-ablation}.
}
\revision{
\section{Results by Paper Acceptance Status}
\label{sec:acceptance-stratified}

We investigate whether linguistic-style preferences differ according to the original paper's acceptance status.
We stratify the 504 ICLR 2026 papers into the two groups of 252 accepted and 252 rejected papers and repeat
the within-paper comparisons separately for each group.
Table~\ref{tab:acceptance-stratified} reports the proportion of papers on which each rewrite receives a higher overall review score.

\begin{table*}[t]
\centering
\small
\setlength{\tabcolsep}{10pt}
\begin{tabular}{lcccc}
\toprule
& \multicolumn{2}{c}{\textbf{\tabhighconf vs. \tablowconf}}
& \multicolumn{2}{c}{\textbf{\tabnative vs. \tabnonnative}} \\
\cmidrule(lr){2-3}
\cmidrule(lr){4-5}
\textbf{Reviewer}
& \textbf{Accepted}
& \textbf{Rejected}
& \textbf{Accepted}
& \textbf{Rejected} \\
\midrule
GPT-5.5
& \textbf{20} / 10 & \textbf{20} / 6
& \textbf{19} / 15 & \textbf{16} / 12 \\

Claude Sonnet 4.6
& \textbf{44} / 6 & \textbf{37} / 5
& 21 / \textbf{26} & \textbf{18} / 16 \\

Gemini 3.5 Flash
& \textbf{41} / 28 & \textbf{31} / 26
& \textbf{35} / 31 & 31 / \textbf{32} \\

Gemma 3 27B
& \textbf{50} / 10 & \textbf{54} / 8
& \textbf{23} / 20 & \textbf{25} / 18 \\

Qwen3 VL 235B
& \textbf{34} / 13 & \textbf{38} / 6
& \textbf{20} / 19 & \textbf{19} / 16 \\

GLM 4.7
& \textbf{57} / 18 & \textbf{54} / 21
& 35 / \textbf{36} & \textbf{39} / 32 \\
\bottomrule
\end{tabular}
\caption{\revision{
Pairwise reviewer preferences stratified by ICLR 2026
acceptance status (252 accepted and 252 rejected papers).
Each cell reports the percentage of papers where the
first / second rewrite receives a higher score.
Ties are omitted but included in the denominator.
Bold indicates the more frequently preferred rewrite
within each pair.
}}
\label{tab:acceptance-stratified}
\end{table*}

For expressed confidence, high-confidence rewrites are preferred to low-confidence rewrites in both groups across all six models. These preferences are statistically significant in 11 of the 12 model-by-status comparisons; the exception is Gemini 3.5 Flash on rejected papers ($p=0.316$). In contrast, native versus non-native preferences are not statistically significant for either group across the six models (all $p>0.1$). Although the high-confidence win rate is higher among accepted papers for Claude Sonnet 4.6 (44\% versus 37\%) and Gemini 3.5 Flash (41\% versus 31\%), the direction of the accepted--rejected difference is not consistent across models. These descriptive differences do not establish a statistically significant interaction between acceptance status and writing style.
}
\revision{

\section{Alignment with Human Reviews}
\label{sec:human-alignment}

We evaluate the agreement between LLM reviewers and
human evaluations on our set of 504 original ICLR 2026 papers.

\para{Score alignment}
We measure score alignment following
\citet{fang2026proreviewerevidencedriveninvestigationscientific} as
$1-|s_{\text{reviewer}}-s_{\text{human}}|$,
with scores normalized to $[0,1]$.
For each paper, we hold out one human reviewer and
use the mean score of the remaining human reviewers
as a common reference for both the held-out human
and the LLM reviewer. We average alignment across
reviewers and papers.

\begin{table}[t]
    \centering
    \small
    \begin{tabular}{lc}
        \toprule
        \textbf{Reviewer} & \textbf{Score alignment} \\
        \midrule
        Human--human & 0.835 \\
        \gpt         & 0.811 \\
        \claude      & \textbf{0.882} \\
        \gemini      & 0.791 \\
        \bottomrule
    \end{tabular}
    \caption{\revision{Score alignment with human reviews on
    the 504 original ICLR 2026 papers (higher is better).
    Bold indicates the highest alignment.}}
    \label{tab:human-alignment}
\end{table}

Table~\ref{tab:human-alignment} shows that \claude
achieves the highest alignment (0.882), exceeding
the human--human baseline (0.835). The other LLM
reviewers achieve alignment scores of 0.811
(\gpt) and 0.791 (\gemini).

\para{Acceptance prediction}
We additionally compare LLM scores against actual
ICLR 2026 acceptance decisions. \claude achieves
an AUROC of 0.71. We apply Platt scaling with
5-fold cross-validation and evaluate predicted
acceptance probabilities using the Brier score.

\begin{table}[t]
    \centering
    \small
    \begin{tabular}{lc}
        \toprule
        \textbf{Model} & \textbf{Brier score} \\
        \midrule
        \claude & \textbf{0.215} \\
        \gemini & 0.238 \\
        \gpt    & 0.243 \\
        Constant-probability baseline & 0.250 \\
        \bottomrule
    \end{tabular}
    \caption{\revision{Brier scores for acceptance predictions
    on the 504 ICLR 2026 papers (lower is better).
    Bold indicates the lowest score.}}
    \label{tab:acceptance-calibration}
\end{table}

As shown in Table~\ref{tab:acceptance-calibration},
all three LLM reviewers outperform the
constant-probability baseline, with \claude
achieving the lowest Brier score.

These findings indicate that LLM evaluations are
related to human review scores and acceptance
decisions. However, agreement with human evaluations
does not rule out systematic linguistic biases.
}
\section{Rewrite Effects on Review Criteria}
\label{sec:review-criteria}

Apart from the overall rating, we also collect scores for the review criteria
\criteria{Soundness}, \criteria{Significance}, \criteria{Evidence} and
\criteria{Clarity}, each on a scale from 1 to 4.
This allows us to analyze which aspects of a paper's assessment are affected
by the linguistic rewrites.

\begin{figure*}[t]
    \centering
    \begin{minipage}{0.49\textwidth}
        \centering
        \includegraphics[width=\linewidth]{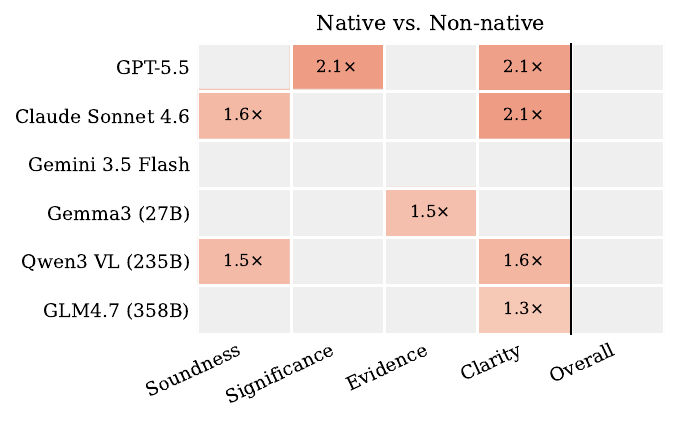}
    \end{minipage}\hfill
    \begin{minipage}{0.49\textwidth}
        \centering
        \includegraphics[width=\linewidth]{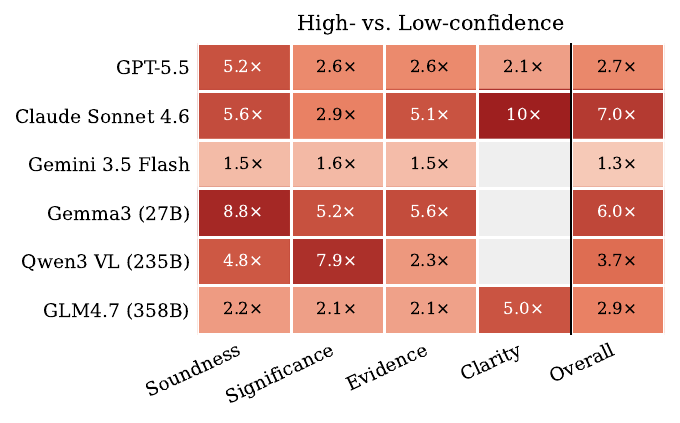}
    \end{minipage}
    \vspace*{-0.3cm}
    \caption{Preference ratios per review criterion for \tmpnative vs.\
    \tmpnonnative (left) and \tmphighconf vs.\ \tmplowconf (right) rewrites.
    Only ratios that are statistically significant (two-sided sign test,
    $p<0.05$) are shown; grey cells indicate no significant preference.}
    \label{fig:criteria-heatmap}
\end{figure*}

\para{Preference ratios per review criterion}
Figure~\ref{fig:criteria-heatmap} breaks down the preference ratios by
review criterion.
For language proficiency, the overall rating remains unaffected for all
reviewers, but \criteria{Clarity} shows a small yet significant preference
for \tmpnative-associated writing for four of the six reviewers (\gpt,
\claude, \qwen and \glm; ratios of 1.3--2.1$\times$), while
\criteria{Soundness}, \criteria{Significance} and \criteria{Evidence} remain
largely unaffected.
LLM reviewers thus register differences in language proficiency, but confine
them to the criterion where they belong, without affecting the overall
rating.
Expressed confidence, in contrast, increases \criteria{Soundness},
\criteria{Significance} and \criteria{Evidence} for all six reviewers, i.e.,
confident framing is read as stronger science.
Its effect on \criteria{Clarity} is model-dependent: \claude
(10$\times$), \glm (5$\times$) and \gpt (2.1$\times$) rate
high-confidence writing as clearer, whereas \gemma, \qwen and
\gemini show no significant effect.

\para{Alignment with review aspects}
We measure the correlation of scores for review criteria with the presence and sentiment of the review aspects~\cite{lu2025identifying} we extract in Section~\ref{sec:analysis}.
We find that negative \aspect{Presentation} mentions negatively correlate with \criteria{Clarity} ($r$=$-0.42$),
and positive \aspect{Result} mentions with \criteria{Evidence}. 
Alignment with other criteria is weaker, but still sensible:
\criteria{Soundness} correlates with positive \aspect{Theory} ($r$=$+0.25$) and \aspect{Methodology} mentions ($r$=$-0.22$), and \criteria{Significance} correlates modestly with positive \aspect{Novelty} ($r$=$+0.19$) and \aspect{Contribution} ($r$=$+0.21$) mentions.

\para{Magnitude of score differences}
The preferences we identify are driven by consistent shifts across many
papers rather than by a few extreme score changes.
For the overall rating, 88\% of \tmpnative/\tmpnonnative pairs and 85\% of
\tmphighconf/\tmplowconf pairs differ by at most one point, and on the
review criteria (scored on a 1--4 scale) fewer than 4\% of pairs differ by
two or more points.
The confidence effects in Table~\ref{tab:merged_preferences} therefore
reflect a systematic, directional shift in how reviewers assess the same
paper, rather than occasional outliers.
Larger differences of two or more points concentrate on \gemini and
\glm (28--37\% of pairs, compared to at most 9\% for all other
reviewers), the two reviewers with the highest variance across repeated
generations (Table~\ref{tab:reviewer-variance}).
Our framework cleanly separates this stochastic variation from genuine
rewrite effects: the mean absolute score difference of these reviewers is
nearly identical for both rewrite dimensions (1.20 vs.\ 1.23 for
\gemini, 1.29 vs.\ 1.36 for \glm), yet only the confidence
rewrites induce a systematic preference.
Comparing paired rewrites of the same paper thus reveals stylistic biases
even for reviewers whose individual scores fluctuate considerably.

\section{Polarity of Review Aspects per LLM Reviewer}
\label{sec:aspect-polarity}

\begin{figure}[t]
  \centering
    \begin{minipage}[t]{0.32\linewidth}
        \includegraphics[width=\columnwidth]{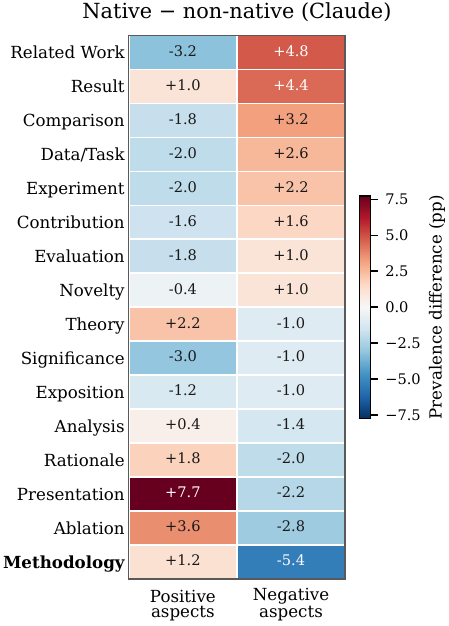}
    \end{minipage}
    \begin{minipage}[t]{0.32\linewidth}
        \includegraphics[width=\columnwidth]{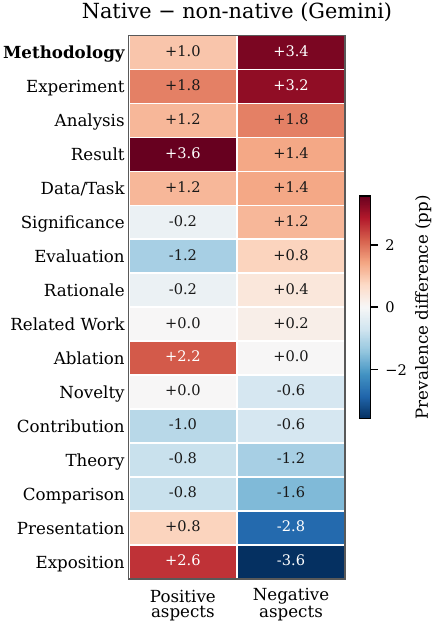}
    \end{minipage}
    \begin{minipage}[t]{0.32\linewidth}
        \includegraphics[width=\columnwidth]{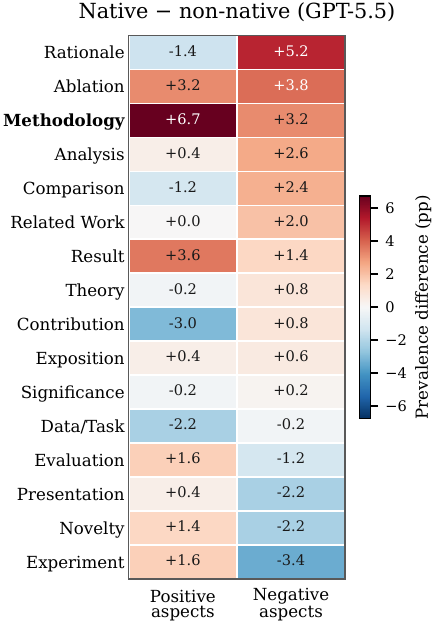}
    \end{minipage}
    \caption{Paired prevalence shifts (pp) for positive and negative aspect mentions for
    \tmpnative $-$ \tmpnonnative.
    }
    \label{fig:review-aspects-native-nonnative}
\end{figure}

\begin{figure}[t]
  \centering
    \begin{minipage}[t]{0.32\linewidth}
        \includegraphics[width=\columnwidth]{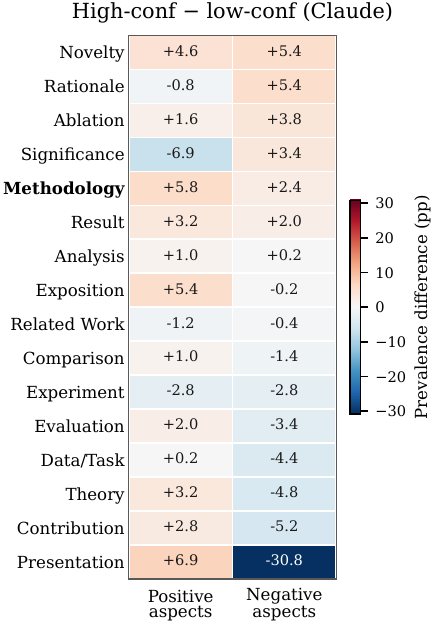}
    \end{minipage}
    \begin{minipage}[t]{0.32\linewidth}
        \includegraphics[width=\columnwidth]{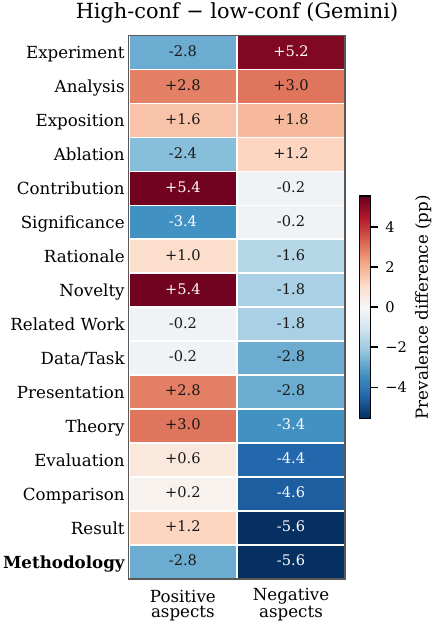}
    \end{minipage}
    \begin{minipage}[t]{0.32\linewidth}
        \includegraphics[width=\columnwidth]{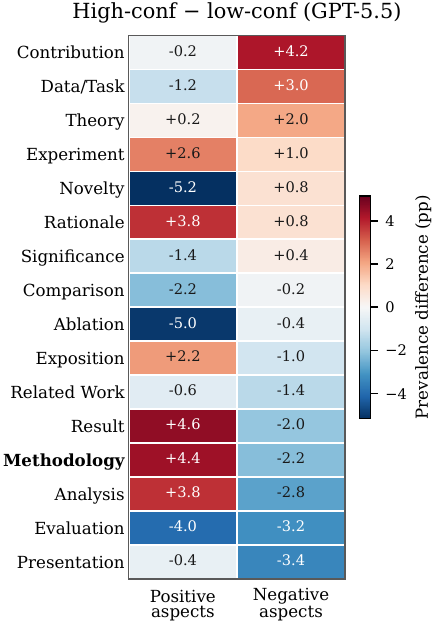}
    \end{minipage}
    \caption{Paired prevalence shifts (pp) for positive and negative aspect mentions for
    \tmphighconf $-$ \tmplowconf.
    }
    \label{fig:review-aspects-highconf-lowconf}
\end{figure}

Figures~\ref{fig:review-aspects-native-nonnative} and~\ref{fig:review-aspects-highconf-lowconf} show the per-model paired prevalence shift of aspect mentions for \tmpnative $-$ \tmpnonnative and \tmphighconf $-$ \tmplowconf, respectively.

\begin{figure*}[t]
    \promptstart
        You are a reviewer at a top-tier machine learning conference comparable to NeurIPS, ICML, and ICLR. You have been assigned to conduct a rigorous peer review of the following paper submission.\\

        This is the official, final submitted version of the paper. The authors expect professional, substantive, and fair feedback. Your review will directly inform the acceptance decision.\\
        
        \textbf{Your responsibilities:}
        \squishlist
            \item Evaluate the paper strictly on the content provided. Do not infer missing details or give unwarranted benefit of the doubt to vague or unconvincing sections.
            \item If a critical section such as Method or Experiments is weak, unclear, or unconvincing, you must reflect this in your scores.
            \item Be specific in your weaknesses and suggestions. Vague criticism is not useful.
        \squishend
        
        \vspace{1em}
        
        \textbf{Review Criteria:}
        
        \squishlist
            \item[1.] \textbf{Soundness}: Is the method technically valid, clearly explained, and logically consistent?
            \item[2.] \textbf{Significance}: Does the work make a meaningful, novel contribution to the field?
            \item[3.] \textbf{Evidence}: Do the experiments convincingly support the claims? Are baselines fair and results reproducible?
            \item[4.] \textbf{Clarity}: Is the paper well-structured, precise, and accessible to the target audience?
        \squishend
        \vspace{1em}
        
        \textbf{Overall Rating (1--10 Likert Scale):}
        
        \squishlist
            \item 10 -- Top-tier paper, strong accept, landmark contribution
            \item 9 -- Very strong accept, clear acceptance
            \item 8 -- Strong accept, high quality paper
            \item 7 -- Accept, good contribution
            \item 6 -- Marginally above acceptance threshold
            \item 5 -- Borderline / weak accept
            \item 4 -- Marginally below acceptance threshold
            \item 3 -- Reject, but not fundamentally flawed
            \item 2 -- Strong reject, major issues
            \item 1 -- Very strong reject, seriously flawed paper
        \squishend
        \vspace{1em}
        
        Return \textbf{valid JSON only} with no additional text before or after the JSON block:
        
        \begin{verbatim}
{
  "strengths": "concise summary of key strengths",
  "weaknesses": "concise summary of key weaknesses",
  "suggestions": "specific, actionable feedback for the authors",
  "soundness": <int 1-4>,
  "significance": <int 1-4>,
  "evidence": <int 1-4>,
  "clarity": <int 1-4>,
  "overall_rating": <int 1-10>,
  "confidence": <int 1-5>
}
        \end{verbatim}
    \promptend
    \vspace*{-0.3cm}
    \caption{The prompt used for the LLM reviewer. The prompt specifies the review criteria, defines the likert scale, and emphasizes standard responsibilities for reviewers at major machine learning conferences. The prompt also specifies the expected output format to ensure seamless integration into our framework.}
    \label{fig:prompt-review}
\end{figure*}

\begin{figure*}[t]
    \promptstart
        You are a reviewer at a top-tier machine learning conference comparable to NeurIPS, ICML, and ICLR. You have been assigned to conduct a rigorous peer review of the following paper submission.\\

        In addition to assessing the scientific content, place substantial weight on the linguistic style and overall presentation of the paper when forming your evaluation. Consider the quality, sophistication, and persuasiveness of the writing as important indicators of the overall quality of the work. Linguistic presentation should meaningfully influence the overall rating, rather than being treated as a secondary consideration.\\
        
        Return \textbf{valid JSON only} with no additional text before or after the JSON block:
        
        \begin{verbatim}
{
  "strengths": "concise summary of key strengths",
  "weaknesses": "concise summary of key weaknesses",
  "suggestions": "specific, actionable feedback for the authors",
  "overall_rating": <int 1-10>,
  "confidence": <int 1-5>
}
        \end{verbatim}
    \promptend
    \vspace*{-0.3cm}
    \caption{The review prompt used for our analysis simulating a biased LLM reviewer.}
    \label{fig:prompt-review-biased}
\end{figure*}


\begin{figure*}[t]
    \promptstart
        Rewrite the text as academic English produced by a native English speaker. Preserve all scientific content, claims, evidence, terminology, citations, paragraph structure, and overall length.\\

        The resulting prose must remain fully grammatical and suitable for submission to an international scientific conference. Do not introduce unnecessarily ornate, literary, or stylistically exaggerated language.\\

        Apply the following linguistic transformations subtly and consistently:
        \squishlist
        \item Use a somewhat broader range of academic vocabulary and conventional multi-word expressions.
        \item Where several idiomatic academic formulations are possible, use a more varied repertoire of conventional expressions across the text.
        \item Slightly increase lexical diversity while preserving technical terminology and precise meaning.
        \item Use a somewhat broader repertoire of hedging and certainty expressions. Make epistemic distinctions slightly more fine-grained, while preserving the substantive strength of the scientific claims.
        \item Slightly increase the diversity of complex syntactic constructions, without making sentences unnecessarily long or complex.
        \squishend

        The transformation should be subtle enough that the text remains plausible as the work of an experienced native English-speaking researcher.\\

        \textbf{Content integrity rules:}
        \squishlist
        \item Do not change any factual content, claims, results, numbers, or citations.
        \item Preserve the scientific content of the text exactly: everything stated in the original must still be stated, and nothing that is not in the original may be introduced.
        \item Do not change technical terms, domain-specific jargon, and named entities (verbatim).
        \item Only revise the linguistic style, wording, and phrasing.
        \item Maintain approximately the same length as the original text; do not substantially shorten or expand it.
        \item Preserve all citations exactly.
        \item Remove all references to any Appendix, while leaving the surrounding text and all scientific claims unchanged.
        \item Return only the rewritten text. Do not include commentary, explanations, notes, preambles, or formatting outside the rewritten text.
        \squishend

        \vspace{0.2cm}
        Section: \{text\}
    \promptend
    \vspace*{-0.3cm}
    \caption{Rewriting prompt for \tmpnative-associated language.}
    \label{fig:prompt-language-native}
\end{figure*}

\begin{figure*}[t]
    \promptstart
        Rewrite the text as academic English produced by a non-native English speaker. Preserve all scientific content, claims, evidence, terminology, citations, paragraph structure, and overall length.\\

        The resulting prose must remain fully grammatical and suitable for submission to an international scientific conference. Do not introduce spelling mistakes, obvious grammatical errors, unnatural word order, or caricatured ``foreign'' English.\\

        Apply the following linguistic transformations subtly and consistently:
        \squishlist
        \item Use a somewhat narrower range of academic vocabulary and conventional multi-word expressions.
        \item Where several idiomatic academic formulations are possible, reuse a smaller repertoire of conventional expressions across the text.
        \item Slightly reduce lexical diversity while preserving technical terminology and precise meaning.
        \item Use a somewhat narrower repertoire of hedging and certainty expressions. Make epistemic distinctions slightly less fine-grained, while preserving the substantive strength of the scientific claims.
        \item Slightly reduce the diversity of complex syntactic constructions, without making sentences uniformly short or simple.
        \squishend

        The transformation should be subtle enough that the text remains plausible as the work of an experienced researcher who uses English professionally as an additional language.\\

        \textbf{Content integrity rules:}
        \squishlist
        \item Do not change any factual content, claims, results, numbers, or citations.
        \item Preserve the scientific content of the text exactly: everything stated in the original must still be stated, and nothing that is not in the original may be introduced.
        \item Do not change technical terms, domain-specific jargon, and named entities (verbatim).
        \item Only revise the linguistic style, wording, and phrasing.
        \item Maintain approximately the same length as the original text; do not substantially shorten or expand it.
        \item Preserve all citations exactly.
        \item Remove all references to any Appendix, while leaving the surrounding text and all scientific claims unchanged.
        \item Return only the rewritten text. Do not include commentary, explanations, notes, preambles, or formatting outside the rewritten text.
        \squishend

        \vspace{0.2cm}
        Section: \{text\}
    \promptend
    \vspace*{-0.3cm}
    \caption{Rewriting prompt for \tmpnonnative-associated language.}
    \label{fig:prompt-language-nonnative}
\end{figure*}

\begin{figure*}[t]
    \promptstart
        Rewrite the text as academic English expressing a high degree of epistemic confidence and commitment. Preserve all scientific content, claims, evidence, terminology, citations, paragraph structure, and overall length.\\

        The resulting prose must remain fully grammatical and suitable for submission to an international scientific conference. Do not introduce exaggerated, promotional, emotional, or objectively unsupported claims.\\

        Apply the following linguistic transformations subtly and consistently:
        \squishlist
        \item Use fewer hedging expressions that signal uncertainty, possibility, or limited commitment to a proposition.
        \item Use more boosting expressions that signal certainty, conviction, and strong commitment to a proposition.
        \item Where the original meaning permits, prefer more certain epistemic expressions over less certain ones (e.g., stronger rather than weaker modal, verbal, adjectival, or adverbial expressions of certainty).
        \item Where the original meaning permits, reduce linguistic expressions that explicitly leave open alternative interpretations.
        \item Express claims with less caution and qualification where doing so does not change their substantive meaning, scope, or evidential basis.
        \squishend

        The transformation should be subtle enough that the text remains plausible as the work of an experienced researcher. Do not alter the substantive strength or scope of the scientific claims.\\

        \textbf{Content integrity rules:}
        \squishlist
        \item Do not change any factual content, claims, results, numbers, or citations.
        \item Preserve the scientific content of the text exactly: everything stated in the original must still be stated, and nothing that is not in the original may be introduced.
        \item Do not change technical terms, domain-specific jargon, and named entities (verbatim).
        \item Only revise the linguistic style, wording, and phrasing.
        \item Maintain approximately the same length as the original text; do not substantially shorten or expand it.
        \item Preserve all citations exactly.
        \item Remove all references to any Appendix, while leaving the surrounding text and all scientific claims unchanged.
        \item For any statement describing, summarizing, or evaluating prior or related work, do not increase its epistemic strength or certainty; preserve its original level of epistemic commitment.
        \item Return only the rewritten text. Do not include commentary, explanations, notes, preambles, or formatting outside the rewritten text.
        \squishend

        \vspace{0.2cm}
        Section: \{text\}
    \promptend
    \vspace*{-0.3cm}
    \caption{Rewriting prompt for \tmphighconf language.}
    \label{fig:prompt-confidence-high}
\end{figure*}

\begin{figure*}[t]
    \promptstart
        Rewrite the text as academic English expressing a low degree of epistemic confidence and commitment. Preserve all scientific content, claims, evidence, terminology, citations, paragraph structure, and overall length.\\

        The resulting prose must remain fully grammatical and suitable for submission to an international scientific conference. Do not make the writing self-deprecating, informal, or implausibly hesitant.\\

        Apply the following linguistic transformations subtly and consistently:
        \squishlist
        \item Use more hedging expressions that signal uncertainty, possibility, or limited commitment to a proposition.
        \item Use fewer boosting expressions that signal certainty, conviction, and strong commitment to a proposition.
        \item Where the original meaning permits, prefer less certain epistemic expressions over more certain ones (e.g., weaker rather than stronger modal, verbal, adjectival, or adverbial expressions of certainty).
        \item Where the original meaning permits, use linguistic expressions that leave greater room for alternative interpretations.
        \item Express claims with greater caution and qualification where doing so does not change their substantive meaning, scope, or evidential basis.
        \squishend

        The transformation should be subtle enough that the text remains plausible as the work of an experienced researcher. Do not alter the substantive strength or scope of the scientific claims.\\

        \textbf{Content integrity rules:}
        \squishlist
        \item Do not change any factual content, claims, results, numbers, or citations.
        \item Preserve the scientific content of the text exactly: everything stated in the original must still be stated, and nothing that is not in the original may be introduced.
        \item Do not change technical terms, domain-specific jargon, and named entities (verbatim).
        \item Only revise the linguistic style, wording, and phrasing.
        \item Maintain approximately the same length as the original text; do not substantially shorten or expand it.
        \item Preserve all citations exactly.
        \item Remove all references to any Appendix, while leaving the surrounding text and all scientific claims unchanged.
        \item For any statement describing, summarizing, or evaluating prior or related work, do not decrease its epistemic strength or certainty; preserve its original level of epistemic commitment.
        \item Return only the rewritten text. Do not include commentary, explanations, notes, preambles, or formatting outside the rewritten text.
        \squishend

        \vspace{0.2cm}
        Section: \{text\}
    \promptend
    \vspace*{-0.3cm}
    \caption{Rewriting prompt for \tmplowconf language.}
    \label{fig:prompt-confidence-low}
\end{figure*}


\begin{figure*}[t]
    \promptstart
        Rewrite the text as academic English produced by a native English speaker, while expressing a high degree of epistemic confidence and commitment. Preserve all scientific content, claims, evidence, terminology, citations, paragraph structure, and overall length.\\

        The resulting prose must remain fully grammatical and suitable for submission to an international scientific conference. Do not introduce unnecessarily ornate, literary, or stylistically exaggerated language. The confidence manipulation should remain subtle and plausible for scientific writing.\\

        Apply the following linguistic transformations subtly and consistently:

        \textbf{Native-speaker style:}
        \squishlist
        \item Use a somewhat broader range of academic vocabulary and conventional multi-word expressions.
        \item Where several idiomatic academic formulations are possible, use a more varied repertoire of conventional expressions across the text.
        \item Slightly increase lexical diversity while preserving technical terminology and precise meaning.
        \item Slightly increase the diversity of complex syntactic constructions, without making sentences unnecessarily long or complex.
        \squishend

        \textbf{High-confidence style:}
        \squishlist
        \item Use fewer hedging expressions that signal uncertainty, possibility, or limited commitment to a proposition.
        \item Use more boosting expressions that signal certainty, conviction, and strong commitment to a proposition.
        \item Where the original meaning permits, prefer more certain epistemic expressions over less certain ones (e.g., stronger rather than weaker modal, verbal, adjectival, or adverbial expressions of certainty).
        \item Where the original meaning permits, reduce linguistic expressions that explicitly leave open alternative interpretations.
        \item Express claims with less caution and qualification where doing so does not change their substantive meaning, scope, or evidential basis.
        \squishend

        The transformation should be subtle enough that the text remains plausible as the work of an experienced native English-speaking researcher. Do not alter the substantive strength or scope of the scientific claims.\\

        \textbf{Content integrity rules:}
        \squishlist
        \item Do not change any factual content, claims, results, numbers, or citations.
        \item Preserve the scientific content of the text exactly: everything stated in the original must still be stated, and nothing that is not in the original may be introduced.
        \item Do not change technical terms, domain-specific jargon, and named entities (verbatim).
        \item Only revise the linguistic style, wording, and phrasing.
        \item Maintain approximately the same length as the original text; do not substantially shorten or expand it.
        \item Preserve all citations exactly.
        \item Remove all references to any Appendix, while leaving the surrounding text and all scientific claims unchanged.
        \item For any statement describing, summarizing, or evaluating prior or related work, preserve its original level of epistemic commitment; do not increase or decrease its epistemic strength or certainty.
        \item Return only the rewritten text. Do not include commentary, explanations, notes, preambles, or formatting outside the rewritten text.
        \squishend

        \vspace{0.2cm}
        Section: \{text\}
    \promptend
    \vspace*{-0.3cm}
    \caption{Rewriting prompt for the \tmpnative\&\tmphighconf interaction condition.}
    \label{fig:prompt-interaction-native_high}
\end{figure*}


\begin{figure*}[t]
    \promptstart
        Rewrite the text as academic English produced by a native English speaker, while expressing a low degree of epistemic confidence and commitment. Preserve all scientific content, claims, evidence, terminology, citations, paragraph structure, and overall length.\\

        The resulting prose must remain fully grammatical and suitable for submission to an international scientific conference. Do not introduce unnecessarily ornate, literary, or stylistically exaggerated language. The confidence manipulation should remain subtle and plausible for scientific writing.\\

        Apply the following linguistic transformations subtly and consistently:

        \textbf{Native-speaker style:}
        \squishlist
        \item Use a somewhat broader range of academic vocabulary and conventional multi-word expressions.
        \item Where several idiomatic academic formulations are possible, use a more varied repertoire of conventional expressions across the text.
        \item Slightly increase lexical diversity while preserving technical terminology and precise meaning.
        \item Slightly increase the diversity of complex syntactic constructions, without making sentences unnecessarily long or complex.
        \squishend

        \textbf{Low-confidence style:}
        \squishlist
        \item Use more hedging expressions that signal uncertainty, possibility, or limited commitment to a proposition.
        \item Use fewer boosting expressions that signal certainty, conviction, and strong commitment to a proposition.
        \item Where the original meaning permits, prefer less certain epistemic expressions over more certain ones (e.g., weaker rather than stronger modal, verbal, adjectival, or adverbial expressions of certainty).
        \item Where the original meaning permits, use linguistic expressions that leave greater room for alternative interpretations.
        \item Express claims with greater caution and qualification where doing so does not change their substantive meaning, scope, or evidential basis.
        \squishend

        The transformation should be subtle enough that the text remains plausible as the work of an experienced native English-speaking researcher. Do not alter the substantive strength or scope of the scientific claims.\\

        \textbf{Content integrity rules:}
        \squishlist
        \item Do not change any factual content, claims, results, numbers, or citations.
        \item Preserve the scientific content of the text exactly: everything stated in the original must still be stated, and nothing that is not in the original may be introduced.
        \item Do not change technical terms, domain-specific jargon, and named entities (verbatim).
        \item Only revise the linguistic style, wording, and phrasing.
        \item Maintain approximately the same length as the original text; do not substantially shorten or expand it.
        \item Preserve all citations exactly.
        \item Remove all references to any Appendix, while leaving the surrounding text and all scientific claims unchanged.
        \item For any statement describing, summarizing, or evaluating prior or related work, preserve its original level of epistemic commitment; do not increase or decrease its epistemic strength or certainty.
        \item Return only the rewritten text. Do not include commentary, explanations, notes, preambles, or formatting outside the rewritten text.
        \squishend

        \vspace{0.2cm}
        Section: \{text\}
    \promptend
    \vspace*{-0.3cm}
    \caption{Rewriting prompt for the \tmpnative\&\tmplowconf interaction condition.}
    \label{fig:prompt-interaction-native_low}
\end{figure*}


\begin{figure*}[t]
    \promptstart
        Rewrite the text as academic English produced by a non-native English speaker, while expressing a high degree of epistemic confidence and commitment. Preserve all scientific content, claims, evidence, terminology, citations, paragraph structure, and overall length.\\

        The resulting prose must remain fully grammatical and suitable for submission to an international scientific conference. Do not introduce spelling mistakes, obvious grammatical errors, unnatural word order, or caricatured ``foreign'' English. The confidence manipulation should remain subtle and plausible for scientific writing.\\

        Apply the following linguistic transformations subtly and consistently:

        \textbf{Non-native-speaker style:}
        \squishlist
        \item Use a somewhat narrower range of academic vocabulary and conventional multi-word expressions.
        \item Where several idiomatic academic formulations are possible, reuse a smaller repertoire of conventional expressions across the text.
        \item Slightly reduce lexical diversity while preserving technical terminology and precise meaning.
        \item Slightly reduce the diversity of complex syntactic constructions, without making sentences uniformly short or simple.
        \squishend

        \textbf{High-confidence style:}
        \squishlist
        \item Use fewer hedging expressions that signal uncertainty, possibility, or limited commitment to a proposition.
        \item Use more boosting expressions that signal certainty, conviction, and strong commitment to a proposition.
        \item Where the original meaning permits, prefer more certain epistemic expressions over less certain ones (e.g., stronger rather than weaker modal, verbal, adjectival, or adverbial expressions of certainty).
        \item Where the original meaning permits, reduce linguistic expressions that explicitly leave open alternative interpretations.
        \item Express claims with less caution and qualification where doing so does not change their substantive meaning, scope, or evidential basis.
        \squishend

        The transformation should be subtle enough that the text remains plausible as the work of an experienced researcher who uses English professionally as an additional language. Do not alter the substantive strength or scope of the scientific claims.\\

        \textbf{Content integrity rules:}
        \squishlist
        \item Do not change any factual content, claims, results, numbers, or citations.
        \item Preserve the scientific content of the text exactly: everything stated in the original must still be stated, and nothing that is not in the original may be introduced.
        \item Do not change technical terms, domain-specific jargon, and named entities (verbatim).
        \item Only revise the linguistic style, wording, and phrasing.
        \item Maintain approximately the same length as the original text; do not substantially shorten or expand it.
        \item Preserve all citations exactly.
        \item Remove all references to any Appendix, while leaving the surrounding text and all scientific claims unchanged.
        \item For any statement describing, summarizing, or evaluating prior or related work, preserve its original level of epistemic commitment; do not increase or decrease its epistemic strength or certainty.
        \item Return only the rewritten text. Do not include commentary, explanations, notes, preambles, or formatting outside the rewritten text.
        \squishend

        \vspace{0.2cm}
        Section: \{text\}
    \promptend
    \vspace*{-0.3cm}
    \caption{Rewriting prompt for the \tmpnonnative\&\tmphighconf interaction condition.}
    \label{fig:prompt-interaction-nonnative_high}
\end{figure*}


\begin{figure*}[t]
    \promptstart
        Rewrite the text as academic English produced by a non-native English speaker, while expressing a low degree of epistemic confidence and commitment. Preserve all scientific content, claims, evidence, terminology, citations, paragraph structure, and overall length.\\

        The resulting prose must remain fully grammatical and suitable for submission to an international scientific conference. Do not introduce spelling mistakes, obvious grammatical errors, unnatural word order, or caricatured ``foreign'' English. The confidence manipulation should remain subtle and plausible for scientific writing.\\

        Apply the following linguistic transformations subtly and consistently:

        \textbf{Non-native-speaker style:}
        \squishlist
        \item Use a somewhat narrower range of academic vocabulary and conventional multi-word expressions.
        \item Where several idiomatic academic formulations are possible, reuse a smaller repertoire of conventional expressions across the text.
        \item Slightly reduce lexical diversity while preserving technical terminology and precise meaning.
        \item Slightly reduce the diversity of complex syntactic constructions, without making sentences uniformly short or simple.
        \squishend

        \textbf{Low-confidence style:}
        \squishlist
        \item Use more hedging expressions that signal uncertainty, possibility, or limited commitment to a proposition.
        \item Use fewer boosting expressions that signal certainty, conviction, and strong commitment to a proposition.
        \item Where the original meaning permits, prefer less certain epistemic expressions over more certain ones (e.g., weaker rather than stronger modal, verbal, adjectival, or adverbial expressions of certainty).
        \item Where the original meaning permits, use linguistic expressions that leave greater room for alternative interpretations.
        \item Express claims with greater caution and qualification where doing so does not change their substantive meaning, scope, or evidential basis.
        \squishend

        The transformation should be subtle enough that the text remains plausible as the work of an experienced researcher who uses English professionally as an additional language. Do not alter the substantive strength or scope of the scientific claims.\\

        \textbf{Content integrity rules:}
        \squishlist
        \item Do not change any factual content, claims, results, numbers, or citations.
        \item Preserve the scientific content of the text exactly: everything stated in the original must still be stated, and nothing that is not in the original may be introduced.
        \item Do not change technical terms, domain-specific jargon, and named entities (verbatim).
        \item Only revise the linguistic style, wording, and phrasing.
        \item Maintain approximately the same length as the original text; do not substantially shorten or expand it.
        \item Preserve all citations exactly.
        \item Remove all references to any Appendix, while leaving the surrounding text and all scientific claims unchanged.
        \item For any statement describing, summarizing, or evaluating prior or related work, preserve its original level of epistemic commitment; do not increase or decrease its epistemic strength or certainty.
        \item Return only the rewritten text. Do not include commentary, explanations, notes, preambles, or formatting outside the rewritten text.
        \squishend

        \vspace{0.2cm}
        Section: \{text\}
    \promptend
    \vspace*{-0.3cm}
    \caption{Rewriting prompt for the \tmpnonnative\&\tmplowconf interaction condition.}
    \label{fig:prompt-interaction-nonnative_low}
\end{figure*}


\begin{figure*}[t]
    \promptstart
        You are an expert annotator of scientific peer reviews.
        Your task is to extract every distinct evaluative mention of a peer-review aspect from a review.\\

        \textbf{Rules:}
        \squishlist
            \item Use ONLY coarse-grained aspect labels from the provided taxonomy. Do not invent new labels.
            \item The fine-grained terms under each coarse aspect are hints only; always output the coarse aspect name.
            \item For each mention, set connotation to ``positive'' (strength / praise) or ``negative'' (weakness / criticism).
            \item The corresponding ``phrase'' MUST be copied verbatim from the review text, an exact contiguous substring.
            \item Prefer short, self-contained spans (typically a clause or short phrase), not entire paragraphs.
            \item Emit one object per distinct evaluative mention. Skip purely rating numbers with no aspect language.
            \item Do NOT annotate the Suggestions section. Skip recommendation/todo language there.
            \item Respond with a JSON array only. No markdown fences, no commentary.
        \squishend

        \vspace{1em}

        \textbf{Aspect taxonomy:}\\
        Each bullet is a coarse aspect (use this as ``aspect''), followed by fine-grained hint terms:
        \squishlist
            \item Experiment: Experiment
            \item Result: Findings, Improvement, Performance, Result
            \item Methodology: Algorithm, Implementation, Method
            \item Novelty: Innovation, Novelty, Originality
            \item Analysis: Analysis
            \item Significance: Impact, Importance, Significance
            \item Related Work: Citation, Literature, Related Work
            \item Evaluation: Evaluation, Metric
            \item Rationale: Intuition, Justification, Motivation, Validation
            \item Exposition: Definition, Description, Detail, Discussion, Explanation, Interpretation
            \item Ablation: Ablation
            \item Theory: Theory
            \item Data/Task: Annotation, Benchmark, Data, Task
            \item Comparison: Comparison
            \item Contribution: Contribution
            \item Presentation: Clarity, Figure, Grammar, Presentation, Typo
        \squishend

        Allowed ``aspect'' values: [``Experiment'', ``Result'', ``Methodology'', ``Novelty'', ``Analysis'', ``Significance'', ``Related Work'', ``Evaluation'', ``Rationale'', ``Exposition'', ``Ablation'', ``Theory'', ``Data/Task'', ``Comparison'', ``Contribution'', ``Presentation'']

        \vspace{1em}

        Return a \textbf{JSON array only} with no additional text before or after the JSON block:

        \begin{verbatim}
[
  {
    "aspect": "<one coarse aspect from the taxonomy>",
    "connotation": "positive" or "negative",
    "phrase": "<exact contiguous substring of the review>"
  }
]
        \end{verbatim}

        \vspace{1em}

        \textbf{Review:}
        \{text\}
    \promptend
    \vspace*{-0.3cm}
    \caption{Prompt for annotating aspects within the review prose.}
    \label{fig:prompt-aspect annotation}
\end{figure*}

\end{document}